\documentclass{article}

\PassOptionsToPackage{numbers, compress}{natbib}

\usepackage[main, final]{neurips_2025}

\usepackage[utf8]{inputenc} 
\usepackage[T1]{fontenc}    
\usepackage{hyperref}       
\usepackage{url}            
\usepackage{booktabs}       
\usepackage{amsfonts}       
\usepackage{nicefrac}       
\usepackage{microtype}      
\usepackage{amsmath}
\usepackage{amssymb}
\usepackage{algorithm}
\usepackage{algorithmic}
\usepackage{xspace}
\usepackage{enumitem}
\usepackage{graphicx}
\usepackage{wrapfig}
\usepackage{multirow}
\usepackage{caption}
\usepackage{tabularx}
\usepackage{subcaption}

\usepackage[dvipsnames,table,xcdraw]{xcolor}

\newcommand{\model}[1][]{\textsc{ConfRover\ifx\\#1\\\else-#1\fi}\xspace}

\newcommand{\specialcell}[2][c]{%
  \begin{tabular}[#1]{@{}c@{}}#2\end{tabular}}

\definecolor{Highlight}{rgb}{0.89,0.89,0.94}

\usepackage[compact]{titlesec}
\usepackage{sidecap}
\titlespacing*{\section}{0pt}{*0.2}{*0.2}
\titlespacing*{\subsection}{0pt}{*0.2}{*0.2}
\titlespacing*{\subsubsection}{0pt}{*0.2}{*0.2}

\newcommand{\bfx}{\mathbf{x}}
\newcommand{\bfz}{\mathbf{z}}
\newcommand{\meanstd}[2]{#1{\tiny $\pm$#2}}

\newcommand{\pluseq}{\mathrel{+}=}
\DeclareMathOperator{\Bin}{Bin}

\DeclareMathOperator{\Linear}{Linear}

\DeclareMathOperator{\OneHot}{OneHot}

\DeclareMathOperator{\TriangleAttentionStartingNode}{TriangleAttentionStartingNode}
\DeclareMathOperator{\TriangleAttentionEndingNode}{TriangleAttentionEndingNode}
\DeclareMathOperator{\TriangleMultiplicationOutgoing}{TriangleMultiplicationOutgoing}
\DeclareMathOperator{\TriangleMultiplicationIncoming}{TriangleMultiplicationIncoming}
\DeclareMathOperator{\PairTransition}{PairTransition}
\DeclareMathOperator{\LayerNorm}{LayerNorm}

\title{Simultaneous Modeling of Protein Conformation and Dynamics via Autoregression}

\author{%
\textbf{Yuning Shen}$^{1}$\textsuperscript{$\star$}, 
\textbf{Lihao Wang}$^{1}$\textsuperscript{$\star$}, 
\textbf{Huizhuo Yuan}$^1$, 
\textbf{Yan Wang}$^{2\dagger}$, \\
\textbf{Bangji Yang}$^{3\dagger}$, 
\textbf{Quanquan Gu}$^{1\ddagger}$\\[1ex]
$^1$ByteDance Seed \\
$^2$School of Mathematical Sciences, Tongji University\\
$^3$Department of Automation, Tsinghua University\\
\texttt{\{yuning.shen,quanquan.gu\}@bytedance.com}}

\begin{document}

\maketitle

\begingroup
\makeatletter
\renewcommand{\@makefnmark}{} 
\renewcommand{\@makefntext}[1]{\parindent 0pt\noindent#1}
\makeatother

\footnotetext{
\noindent
$^{\star}$Equal contribution. 
$^\dagger$Work done during their internship at ByteDance.
$^\ddagger$Corresponding Author.
}
\endgroup

\begin{abstract}
Understanding protein dynamics is critical for elucidating their biological functions. 
The increasing availability of molecular dynamics (MD) data enables the training of deep generative models to efficiently explore the conformational space of proteins.
However, existing approaches either fail to explicitly capture the temporal dependencies between conformations or do not support direct generation of time-independent samples.
To address these limitations, we introduce \model, an autoregressive model that simultaneously learns protein conformation and dynamics from MD trajectories, supporting both time-dependent and time-independent sampling.
At the core of our model is a modular architecture comprising: (i) an \textit{encoding layer}, adapted from protein folding models, that embeds protein-specific information and conformation at each time frame into a latent space; (ii) a \textit{temporal module}, a sequence model that captures conformational dynamics across frames; and (iii) an SE(3) diffusion model as the \textit{structure decoder}, generating conformations in continuous space.
Experiments on ATLAS, a large-scale protein MD dataset of diverse structures, demonstrate the effectiveness of our model in learning conformational dynamics and supporting a wide range of downstream tasks. 
\model is the first model to sample both protein conformations and trajectories within a single framework, offering a novel and flexible approach for learning from protein MD data. Project website: \url{https://bytedance-seed.github.io/ConfRover}.
\end{abstract}

\section{Introduction}

Proteins are flexible molecules that can adopt multiple structures, called \textit{conformations}. 
Their ability to transition between different conformations enables biological processes critical to life.
Characterizing the behavior of a protein, including its \textit{(1) dynamic motions, (2) conformational distribution, and (3) transitions between different states}, is crucial for understanding its function and guiding the design of novel proteins~\citep{berendsen2000collectivedynamics,frauenfelder1991energydynamics,mccammon1984proteindynamics}.
Molecular dynamics (MD) simulations are the ``gold standard'' for studying protein conformational changes~\citep{Childers2017_mddesign,karplus2005molecularmd,mccammon1977dynamicsmdbpti}. These simulations use physical models to describe the energy of a protein conformation and the forces acting on its atoms. By simulating atomic motion through classical mechanics and iteratively sampling conformations over time, MD enables researchers to explore different conformations, approximating the conformational distribution at equilibrium, and gain mechanistic insights in protein functions.
However, MD simulations are both computationally expensive and technically challenging due to long simulation times and the tendency to become trapped in local energy minima.

\begin{figure}[tb]
    \centering
    \includegraphics[width=1.0\linewidth]{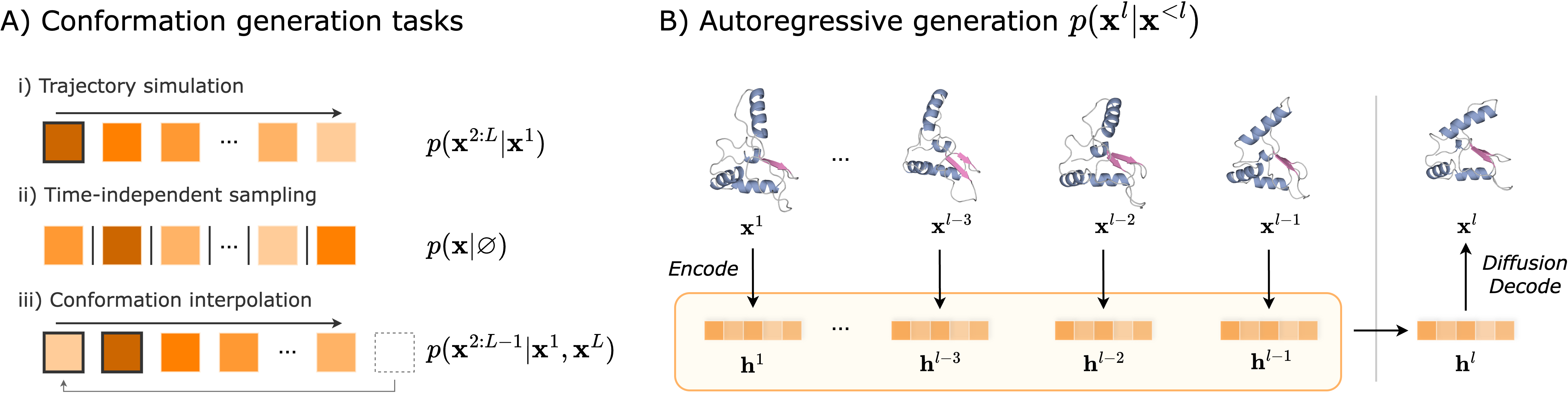}
    \caption{Key ideas of \model. (A) Conformation generation tasks with various conditioning configurations. Each block denotes a frame and arrows indicates the sequential dependencies among frames from autoregressive formulation. Initial conditioning frames are outlined in black. In conformation interpolation, the last frame is repositioned and prepended to the first frame for proper sequential dependencies.
    (B) \model models each frame as a conditional distribution given preceding frames. Sequential dependencies are captured through latent variables $\mathbf{h}$, and conformations are sampled from a diffusion decoder, conditioned on the updated latent.}
    \label{fig:toc}
    \vspace{-6pt}
\end{figure}

These challenges have motivated the use of deep generative models to study proteins, leveraging the rich conformational and dynamic information provided by large-scale MD datasets~\citep{liuDynamicPDBNew2024,yannATLASProteinFlexibility2024}:

(1) \textit{Generating the dynamic motions of proteins is a direct analog to MD simulation.} Pioneering works modeled this by learning transition probabilities of future conformations from the current state~\citep{costa2024equijump,klein2024timewarp,schreiner2024implicit}. However, MD trajectories are often non-Markovian due to partially observed coordinates (e.g., protein-only atoms) and environmental coupling (e.g., using Langevin thermostats). To mitigate this, \citet{cheng20244d4ddiff} incorporated higher-order information using multiple context frames, though this requires fix context windows and limits flexibility. \citet{jing2024generative_mdgen} instead modeled the joint distribution over the entire trajectory, capturing complex dependencies among frames. Due to training on fixed-length trajectories and the non-autoregressive design, their model has limited inference-time flexibility that cannot generate variable-length trajectories. \citet{li2025geometricGST} introduced an autoregressive approach for flexible trajectory extension, but its deterministic formulation cannot capture trajectory distributions or generate diverse samples.

(2) \textit{Learning the conformational distribution enables sampling time-independent conformations.} Several methods~\citep{jingAlphaFoldMeetsFlow2024,lewis2024scalableBioemu,noe2019boltzmann,wang2024proteinconfdiff,zhengPredictingEquilibriumDistributions2024aDiG} train diffusion- or flow-based generative models on conformation ensembles from MD simulation data, bypassing the need for sequential sampling. While effective for generating samples in parallel, they disregard temporal information in MD trajectories and therefore cannot simulate physical motion of proteins.

(3) \textit{Conformation interpolation generate transition pathways between different states.} Recent works~\citep{du2024doobs,jing2024generative_mdgen} have extended generative modeling to conformation interpolation, where the goal is to generate plausible intermediate samples between known start and end conformational states. \citet{jing2024generative_mdgen} framed interpolation as a conditional trajectory generation task, but it requires training task-specific model and has not been evaluated on large proteins.

As these generative problems all stem from the same underlying physical principles and  involve sampling from a protein's conformational space, a natural question arises: \textit{can we develop a general framework to learn all of these objectives?}

We present \model, a framework for simultaneous learning protein conformation distribution and dynamics from MD trajectory data (Figure~\ref{fig:toc}). 
Our key observation is that, for an MD trajectory $\mathbf{x}^{1:L}$ of length $L$, by adopting a general autoregressive formulation, $p(\mathbf{x}^{1:L}) =\prod_{l=1}^L p(\mathbf{x}^l|\mathbf{x}^{<l})$, where $\mathbf{x}^{<l}$ denotes all preceding frames of $\mathbf{x}^l$ in the sequence, we can unify multiple generation objectives as instances of frame generation: (1) generating future frame conditioned on all previous frames, suitable for simulating non-Markovian dynamics; (2) unconditional single-frame generation, $p(\mathbf{x}|\varnothing)$, corresponding to time-independent conformation sampling; (3) flexible frame sequence ordering redefines the dependency structure, enabling tasks such as conformation interpolation.

Our contributions are summarized as follows:
\begin{itemize}[leftmargin=*,labelindent=0pt]
\vspace{-10pt}
\setlength\itemsep{-0.5em}
\item We introduce a simple yet general framework to learn both the conformational distribution and dynamics from MD data, supporting multiple generation tasks including trajectory simulation, time-independent conformation sampling, and conformation interpolation.
\item  We design a modular architecture that captures temporal dependencies in latent space using efficient causal transformers (i.e., Llama~\citep{touvron2023llama}), and directly models conformations in continuous SE(3) space using a diffusion decoder, avoiding discretization structure into tokens.
\item Experiments show strong capabilities of \model: it outperforms \textsc{MDGen}~\citep{jing2024generative_mdgen} in trajectory simulation, matches the performance of \textsc{AlphaFlow}\citep{jingAlphaFoldMeetsFlow2024} and \textsc{ConfDiff}\citep{wang2024proteinconfdiff} in time-dependent generation, and can effectively sample conformations interpolating two endpoints.
\end{itemize}

\section{Background}

\subsection{Data Generation from Molecular Dynamics}

Molecular dynamics describes the motion of molecules through Newtonian mechanics $M \ddot{\mathbf{x}} = -\nabla U(\mathbf{x})$, where $\mathbf{x}$ denotes coordinates of atoms in the system, $M$ is the atomic mass, $U(\mathbf{x})$ is the potential energy of the configuration and $-\nabla U(\mathbf{x})$ represents the forces acting on atoms. In practice, stochastic and frictional forces are integrated to model energy exchange with the environment and maintain temperature control of the system, converting the equations of motion to a Langevin process:
\begin{align*}
    M \ddot{\mathbf{x}} = -\nabla U(\mathbf{x}) - \gamma M \dot{\mathbf{x}} + \sqrt{2M\gamma k_{B}T} \boldsymbol{\eta}(t),
\end{align*}
where $\gamma$ is the friction coefficient, $k_B$ is the Boltzmann constant, $T$ is the temperature, and $\boldsymbol{\eta}(t)$ is a Gaussian noise term delta-correlated in time $\langle \eta_i(t)\eta_j(t')\rangle = \delta_{ij}\delta(t - t')$.
Sampling from this stochastic process generates a time evolution of system configurations. Over time, the samples converges to the Boltzmann distribution $p(\mathbf{x})\propto \exp(-U(\mathbf{x})/k_BT)$.
The trajectory of protein coordinates, $\mathbf{x}^{1:L}_\text{prot} = (\mathbf{x}_\text{prot}^1, \mathbf{x}_\text{prot}^2, \dots, \mathbf{x}_\text{prot}^L)$, is extracted and saved at prescribed simulation intervals, providing both distributional and kinetic information in the protein conformational dynamics. 
For simplicity, we omit the subscript `prot' and use $\mathbf{x}$ to denote protein coordinates throughout the paper.

\subsection{Protein Representations}
\label{sec:prot_rep}

Proteins are chain-like molecules composed of amino acid \textit{residues}, each selected from 20 standard amino acid types. 
We parameterize the coordinates of heavy atoms in a protein using the SE(3)-torsional convention~\citep{jumperHighlyAccurateProtein2021AF2}: the backbone atoms (\texttt{N-C$\alpha$-C}) of each residue define a local coordinate via a Gram-Schmidt process, referred to as a \textit{rigid}. The position and orientation of each rigid relative to the global coordinate system are described by a translation-rotation transformation in SE(3) space.
The backbone conformation of a protein with $N$ residues can then be represented as $\mathbf{x}=(\mathbf{T}, \mathbf{R}) \in \mathrm{SE(3)}^N$, where $\mathbf{T}\in \mathbb{R}^{N\times 3}$ and $\mathbf{R} \in \mathrm{SO(3)}^N$ are the translation and rotation components. 
The coordinates of the oxygen atom of the backbone and the side chain atoms can be determined with the addition of up to 7 torsional angles $(\phi, \psi, \omega, \chi_1, \dots, \chi_4)$ describing the bond rotation. Therefore, the complete configuration of a protein structure is parameterized in the space:
$
\mathbf{x} = (\mathbf{T}, \mathbf{R}, \phi, \psi, \omega, \chi_1, \dots, \chi_4) \in (\mathrm{SE(3)} \times \mathbb{T}^7)^N.
$

\subsection{SE(3)-Diffusion Models for Protein Conformation Generation}
\label{sec:se3}

Diffusion generative models are capable of learning complex data distributions. 
Training involves progressively corrupting data with noise and learning to reverse this process through denoising, thereby modeling the original data distribution~\citep{hoDenoisingDiffusionProbabilistic2020ddpm,songScoreBasedGenerativeModeling2021}.
Recently, diffusion models operating in SE(3) space have been proposed to model protein backbone structures~\citep{wang2024proteinconfdiff,yimSEDiffusionModel2023}. Below, we provide a brief overview of diffusion model and defer the detailed SE(3) formulation to Appendix~\ref{ap:se3_diff}:

Given a protein backbone conformation as $\mathbf{x}_0 = (\mathbf{T}_0, \mathbf{R}_0) \in \mathrm{SE(3)}^N$, 
 and conditioned on the protein identity (omitted in the equations for clarity), we aim to train a neural network to jointly estimate the score functions of the reverse-time marginal distributions at varying diffusion time $t$, 
$s_{\theta}(\mathbf{x}_t, t)\approx \nabla_{\mathbf{x}_t}\log p_{t}(\mathbf{x}_t)$. This model is trained
using the \textit{denoising score matching} (DSM) loss:
\begin{align}
\mathcal{L}_{\text{DSM}} = \mathbb{E}_{\mathbf{x}_0, \mathbf{x}_t, t}\left[\lambda(t) \| s_\theta(\mathbf{x}_t, t) - \nabla_{\mathbf{x}_t} \log p_{t|0}(\mathbf{x}_t | \mathbf{x}_0) \|^2\right] \nonumber. 
\end{align}
Here $p_{t|0}(\mathbf{x}_t|\mathbf{x}_0)$ is the forward transition kernel defined in the SE(3) space, $\mathbf{x}_t = (\mathbf{T}_t, \mathbf{R}_t)$ is the noisy data at time $t$, and $\lambda(t)$ is a time-dependent weight.
During inference, DPM generates clean conformations from random noise by simulating the reverse diffusion process with the learned score network $s_\theta(\mathbf{x}_t, t)$.

\section{\textsc{ConfRover}}
\label{sec:mtd_all}

\subsection{Modeling MD Trajectories through Autoregression}
\label{sec:mtd_auto}
Autoregressive generative models factorize the distribution of a sequence as a series of conditional generations over frames. We cast this idea to MD trajectories, modeling a sequence of $L$ frames as:
\begin{align}
\label{eq:traj_decomp}
p(\mathbf{x}^{1:L}|\mathcal{P})= \prod_{l=1}^{L} p(\mathbf{x}^l|\mathbf{x}^{<l}, \mathcal{P}),
\end{align}
where $\mathbf{x}^{<l}$ is the preceding frames and $\mathcal{P}$ denotes the protein-specific conditioning input.

Despite its simplicity, this formulation naturally supports multiple learning objectives in protein conformation modeling. In its base form, it models temporal dependencies among frames by learning to generate the trajectory. When $L=1$, it removes the frame context and reduces to a single-frame distribution $p(\mathbf{x}|\mathcal{P})$, learning to direct sample time-independent conformations.
In addition, the sequential dependency in Equation~\eqref{eq:traj_decomp} can be extended to any desired frame-conditioned generation tasks. By prepending conditioning frames $\mathcal{K}$ to the sequence, we train the model to learn any conditional generation $p(\mathbf{x}^{1:L}|\mathcal{K},\mathcal{P})$, including conformation interpolation by setting $\mathcal{K}=\{\mathbf{x}^1, \mathbf{x}^L\}$. A similar idea was applied in text infilling tasks by shuffling the order of text contexts~\cite{bavarian2022efficienttraininglanguagemodelsInfill}.

After defining the main learning objectives for trajectory simulation, single-frame (time-independent), and conformation interpolation, we describe how to effectively model autoregressive dependencies over protein conformations in Section~\ref{sec:latent_causal_modeling}. More critically, in Section~\ref{sec:training}, we explain how to adapt sequence models, traditionally designed for discrete tokens, to the continuous space of protein conformations. Lastly, we introduce a specific choice of architectures of \model in Section~\ref{sec:model_arch}.

\subsection{Latent Causal Modeling}
\label{sec:latent_causal_modeling}

\begin{wrapfigure}{r}{0.3\textwidth}  
    \vspace{-12pt}  
    \centering
    \includegraphics[width=0.3\textwidth]{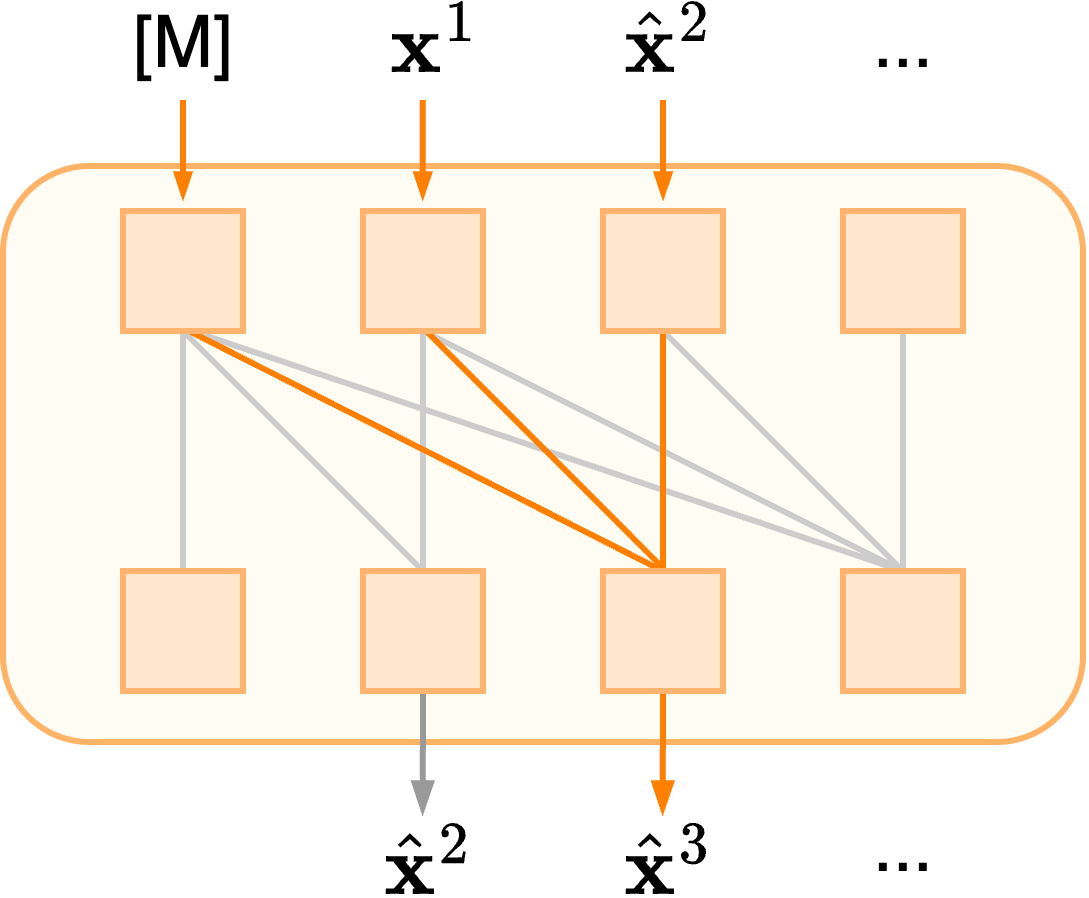}
    \caption{Causal sequence model to generate trajectory ($\hat{\mathbf{x}}^2, \dots$) from the mask token ``[M]'' and the conditioning frame $\mathbf{x}^1$. Each frame only attend to its previous frames. Attention activations for $\hat{\mathbf{x}}^3$ are highlighted in orange.}
    \label{fig:seq_causal}
    \vspace{16pt}  
\end{wrapfigure}

We propose a modular design composed of an encoder, a latent sequence model, and a stochastic decoder. This design enables the use of modern causal transformers, such as in Llama \citep{touvron2023llama}, to efficiently capture sequential dependencies between frames in the latent space (Figure \ref{fig:seq_causal}). During training, the input sequence is shifted by one frame and a mask token ``[M]'' is prepended; the generation process also begins with the mask token and conditioning frames (e.g., $\mathbf{x}^1$). 

To model $p(\mathbf{x}^l|\mathbf{x}^{<l}, \mathcal{P})$, the context frames $\mathbf{x}^{<l}$ are first encoded into intermediate latent states $\mathbf{h}^{<l}=(\mathbf{h}^1, \dots, \mathbf{h}^{l-1})$ using a shared encoder network with protein-specific condition $\mathcal{P}$:
\begin{align}
\label{eq:enc}
\mathbf{h}^i &= f_\eta^\text{enc}(\mathbf{x}^i, \mathcal{P}), \quad i=1, 2, \dots, l-1.
\end{align}
A temporal module is then used to capture the sequential dependencies between the frame latents. Causal attention is used to ensure the latent for frame $\mathbf{h}^{l}$ is only updated by its preceding frames:
\begin{align}
    \label{eq:latent_seq}
    \mathbf{h}^{l}_{\text{updated}} &= f_\xi^\text{temp} \left(\mathbf{h}^1, \mathbf{h}^2, \dots, \mathbf{h}^{l-1}\right).
\end{align}
Since both the encoder and the temporal module are chosen to be deterministic, $p(\mathbf{x}^l|\mathbf{x}^{<l})$ reduces to a conditional generation over the updated latent, realized through a probabilistic decoder:
\begin{align}
\label{eq:dec}
p(\mathbf{x}^l|\mathbf{x}^{<l}, \mathcal{P}) &= p_\theta^{\text{dec}}(\mathbf{x}^{l}|\mathbf{h}^{l}_\text{updated}).
\end{align}
Details on this latent modeling are provided in Appendix \ref{ap:latent_factorize}.
As a result, we train a model $(\eta, \xi, \theta)$ by jointly optimizing the three modules $f_\eta^\text{enc}$, $f_\xi^\text{temp}$, and $p_\theta^{\text{dec}}$.

This setup easily accommodates learning single-frame distribution: by replacing all input frames with a mask token and using identity attentions, where each frame only attends to itself, we effectively disable inter-frame information flow. This trains the model to directly sample conformations. A similar strategy has been used in image-video training~\citep{ho_video_2022,liu2024mardini}.

\subsection{Training Autoregressive Model with SE(3) Diffusion Loss}
\label{sec:training}

The main challenge in applying autoregressive modeling to conformation trajectories lies in representing the continuous distribution of protein conformations within a framework typically used for discrete token sequences. 
While some studies have approached this by discretizing the protein structural space into discrete tokens~\citep{hayesSimulating500Million,liu_diffusion_2023pvqd,lu2024structureesmdiff}, such approaches inherently suffer from discretization error, which can lead to suboptimal performance in modeling protein conformations.

Instead, we propose to directly model the continuous conformational space using diffusion probabilistic models and employ the DSM loss for autoregressive model training, similar to~\citet{li2024autoregressivemar}. Specifically, we perform DSM loss training in SE(3) space.

Given a clean frame $\mathbf{x}^l_0=(\mathbf{T}_0^l, \mathbf{R}_0^l)$, its latent embedding with temporal context $\mathbf{h}^l$ (omit subscript ``update'' for clarity), the forward transition kernels $p_{t|0}(\mathbf{T}_t^l | \mathbf{T}_0^l)$ and $p_{t|0}(\mathbf{R}_t^l | \mathbf{R}_0^l)$ for the translation and rotation component of SE(3), and a score network to jointly estimate the translation and rotation scores $s_\theta(\mathbf{T}_t^l,\mathbf{h}^l, t)$ and $s_\theta^r(\mathbf{R}_t^l,\mathbf{h}^l, t)$, the loss is defined as

\begin{align}
        \mathcal{L}_\text{DSM}^{\text{SE(3)}} &=
\mathbb{E} \left[ \lambda(t) \| s_\theta(\mathbf{T}_t^l,\mathbf{h}^l, t) - \nabla_{\mathbf{T}_t^l} \log p_{t|0}(\mathbf{T}_t^l | \mathbf{T}_0^l) \|^2\right] \nonumber\\ 
&\quad\quad+ \mathbb{E} \left[
\lambda^r(t) \| s_\theta^r(\mathbf{R}_t^l,\mathbf{h}^l, t) - \nabla_{\mathbf{R}_t^l} \log p_{t|0}(\mathbf{R}_t^l | \mathbf{R}_0^l) \|^2 \right]
,
\label{eq:dsm_se3}
\end{align}
where the expectation is taken over the diffusion time $t$ and noisy structure $\mathbf{x}_t^l=(\mathbf{T}_t^l, \mathbf{R}_t^l)$ sampled from the forward process.
Gradients with respect to $\mathbf{h}^l$ are then backpropagated to update the weights in the temporal module $f_\xi^\text{temp}$ and encoder $f_\eta^\text{enc}$.
During inference, we decode each frame autoregressively by performing reverse sampling as in Equation~\eqref{eq:rev_diff}, replacing the scores with estimated values from $s_\theta(\mathbf{T}_t^l,\mathbf{h}^l, t)$ and $s_\theta^r(\mathbf{R}_t^l,\mathbf{h}^l, t)$.

\subsection{Model Architecture}
\label{sec:model_arch}
An overview of model architecture is shown in Figure~\ref{fig:model_arch}, with detailed illustrations of each module provided in Appendix~\ref{ap:arch_details}.

\begin{figure}[t]
    \centering
    \includegraphics[width=\linewidth]{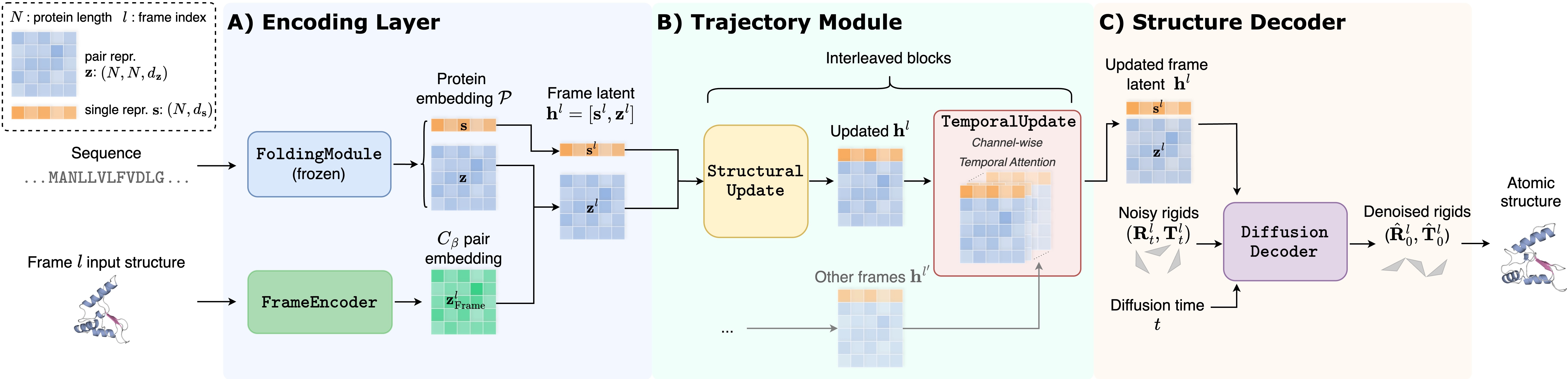}
    \caption{Architecture overview. (A) \textit{Encoding Layer} embeds protein sequence and input structure to each frame as a frame latent representation $\mathbf{h}^l$, comprised of single and pair embeddings; (B) The \textit{Trajectory Module} then updates frame latent $\mathbf{h}^l$ using interleaved structural and temporal update blocks; (C) A diffusion-based \textit{Structure Decoder} learns to denoise noisy conformations conditioned on the updated frame latent $\mathbf{h}^l$; during inference, it samples conformations from the prior distribution. See Appendix \ref{ap:arch_details} for details.}
    \label{fig:model_arch}
\end{figure}


\noindent\textbf{Encoding Layer.}
\label{sec:encoder}
A \texttt{FoldingModule}, parameterized by a pretrained \textsc{OpenFold} model~\citep{ahdritzOpenFoldRetrainingAlphaFold22023}, extracts protein-specific embeddings $\mathcal{P}$ consisting of a single representation ($\mathbf{s}$) and a pair representation ($\mathbf{z}$), shared across frames. For each frame, a \texttt{FrameEncoder}, adapted from the template module used in prior works~\citep{jingAlphaFoldMeetsFlow2024,jumperHighlyAccurateProtein2021AF2}, encodes pairwise distance of pseudo–C$_\beta$ atoms via triangular updates and merges this frame pair representations $\mathbf{z}_{\text{Frame}}^l$ with the protein pair representation $\mathbf{z}$. The resulting frame latent embedding, $\mathbf{h}^l=[\mathbf{s}^l, \mathbf{z}^l]$, is invariant to global translation and rotation of the conformations, and are passed to the \textit{Trajectory Module}.
Following causal sequence modeling, a masked frame token ``[M]'' is introduced by zeroing out the pseudo-C$_\beta$ pairwise distances to remove structural information.

\noindent\textbf{Trajectory Module.}
\label{sec:temp}
The \textit{Trajectory Module} models structural and temporal dependencies across frames, updating each frame's embedding based on its preceding frames. It consists of interleaved \texttt{StructuralUpdate} and \texttt{TemporalUpdate} layers that operate on the frame-wise latent embeddings.
\texttt{StructuralUpdate} incorporates \textit{Pairformer} layers, a core architecture in protein structure modeling, to update single and pair embeddings through triangular operations~\citep{ahdritzOpenFoldRetrainingAlphaFold22023}.
\texttt{TemporalUpdate} employs a Llama-based causal transformer layer for channel-wise self-attention over the sequence of frame embeddings. Each channel in the single and pair embeddings is updated independently. Frame indices are encoded using Rotary Position Embedding \citep{su2023roformerenhancedtransformerrotary}.
This interleaved design enables efficient updates while maintaining flexibility in modeling sequential dependencies.

\noindent\textbf{Structure Decoder.}
\label{sec:struct_decoder}
The updated latent embeddings from \textit{Trajectory Module} serve as conditioning signals for generating the conformation at each frame. 
For the SE(3) diffusion model described in Section~\ref{sec:se3}, we adopt \textsc{ConfDiff}~\citep{wang2024proteinconfdiff} as the \texttt{DiffusionDecoder} to generate 3D conformations. \textsc{ConfDiff} composes of layers of Invariant Point Attention and Transformer (on single embeddings) to collectively update the residue SE(3) \textit{rigids}, as well as single and pair embeddings. 
Trained with denoising score matching in Equation~\eqref{eq:dsm_se3}, the \texttt{DiffusionDecoder} learns to iteratively denoise noisy frame structures drawn from a prior SE(3) distribution, conditioned on the frame latent embeddings, to generate accurate backbone conformations of the frame.
To reconstruct full-atom geometry, we additionally predict the 7 torsional angles $(\phi, \psi, \omega, \chi_1, \dots, \chi_4)$ using a light-weight \texttt{AngleResNet}, for the coordinates of backbone oxygen atom and side-chain atoms.

\section{Experiments}
\label{sec:exp_all}

\noindent\textbf{Dataset.} We evaluate model performance on ATLAS~\citep{yannATLASProteinFlexibility2024}, a large-scale protein MD dataset covering $\sim$1300 proteins with diverse sizes and structures. For each protein, it contains triplicated 100~ns simulation trajectories. All models are trained on training trajectories and evaluated on test trajectories split by protein identity~\citep{jingAlphaFoldMeetsFlow2024,jing2024generative_mdgen,wang2024proteinconfdiff}. This presents a challenging task for assessing the generalization to unseen protein structures and dynamics.

\noindent\textbf{Model training.} 
We use \textsc{OpenFold} (with frozen pretrained weights) as \model's \texttt{FoldingModule} and initialized the weights of the \texttt{DiffusionDecoder} from \textsc{ConfDiff}, while training the remaining parts of the network from scratch.
During training, trajectories of length $L=8$ with varying timesteps (strides), corresponding to $1\sim 1024$ MD snapshots saved at 10~ps intervals, are sampled to enable learning across multiple timescales.
For the base \model, we adopt a \textit{hybrid training} strategy with 1:1 ratio between trajectory and single-frame training objectives. To further enable conformation interpolation, we continue training the base model with a 1:1:1 ratio of trajectory, single-frame and interpolation objectives, denotes the model as \model[interp]. See Appendix~\ref{ap:training_details} for training details.

\noindent\textbf{Baselines.} 
We compare \model with state-of-the-art deep learning models for each task: For \textit{trajectory simulation}, we compare against \textsc{MDGen}~\citep{jing2024generative_mdgen}, a flow-based non-autoregressive trajectory model trained on ATLAS; For \textit{time-independent generation}, we evaluate against \textsc{AlphaFlow}~\citep{jingAlphaFoldMeetsFlow2024} and \textsc{ConfDiff}~\citep{wang2024proteinconfdiff}, flow- and diffusion-based conformation generation models finetuned on ATLAS; For \textit{conformation interpolation}, no existing baseline is available for large proteins, therefore, we focus on analyzing results of our model. See Appendix~\ref{ap:discuss} for more details on the availability of baseline models.

\subsection{Trajectory Simulation}

Since trajectories from both MD and models are stochastic samples, directly comparing them using frame-wise error, such as root-mean-square-deviation (RMSD) between atomic coordinates, is not appropriate. Therefore, we evaluate the model's ability to recover trajectory dynamics from two perspectives: (1) how well it captures the magnitude of conformation changes across varying start conformations and timescales; (2) how well it recovers the conformational states and principal dynamic modes observed in long-time MD simulations.

\noindent\textbf{Evaluating conformation change accuracy on \textit{multi-start} benchmark.} We curated a test benchmark consisting of short trajectories with $L = 9$ frames, extracted from 82 ATLAS test proteins. For each protein, we choose from varying starting frames (snapshot index 1000 $\sim$ 7000) and strides (128 $\sim$ 1024 snapshots), resulting in a total of $\sim2,700$ generation conditions. 
For each trajectory, we measure three aspects of conformational changes: \textit{Trajectory}, the total changes over the entire sequence $\sum_{l=1}^{L-1} d(\mathbf{x}^l, \mathbf{x}^{l+1})$; \textit{Frame}, the changes of each frame relative to the starting frame $d(\mathbf{x}^l, \mathbf{x}^1)$; \textit{$\Delta$~Frame}, the changes between consecutive frames $d(\mathbf{x}^l, \mathbf{x}^{l+1})$. Here, $d(\cdot,\cdot)$ measures the distance between two conformations. We report both the $L^2$-distance in projected 2D PCA space and the RMSD (in Å) of alpha-carbon (C$\alpha$) atoms.
This benchmark captures diverse dynamics at both the trajectory and frame levels and enables comprehensive evaluation across varying conditions and timescales (see Appendix \ref{ap:multi-start} for details).

\begin{wraptable}{r}{0.5\textwidth}
    \vspace{-6pt}
    \centering
    \captionof{table}{Pearson correlations between conformation changes in model-generated and reference trajectories under the \textit{multi-start} setting. The mean and standard deviations are calculated from five independent runs. Models with higher correlations are highlighted in \textbf{bold}.}
    \label{tab:fwd_vary_corr}
    {\footnotesize
    \begin{tabularx}{0.5\textwidth}{cccc}
        \toprule
        & \multicolumn{3}{c}{C$\alpha$ coordinates} \\
        \cmidrule{2-4}
        & Traj. & Frame & $\Delta$Frame \\
        \midrule
        \textsc{MDGen} & \meanstd{0.56}{0.03} & \meanstd{0.47}{0.03} & \meanstd{0.41}{0.02} \\
        \model & \textbf{\meanstd{0.75}{0.01}} & \textbf{\meanstd{0.63}{0.01}} & \textbf{\meanstd{0.53}{0.01}} \\
        \midrule
        & \multicolumn{3}{c}{PCA 2D} \\
        \cmidrule{2-4}
        & Traj. & Frame & $\Delta$Frame \\
        \midrule
        \textsc{MDGen} & \meanstd{0.18}{0.01} & \meanstd{0.15}{0.01} & \meanstd{0.10}{0.01} \\
        \model & \textbf{\meanstd{0.73}{0.01}} & \textbf{\meanstd{0.50}{0.01}} & \textbf{\meanstd{0.43}{0.00}} \\
        \bottomrule
    \end{tabularx}
    }
    \vspace{-12pt}
\end{wraptable}

\begin{figure}[t]
    \centering
    \centering
    \includegraphics[width=0.95\linewidth]{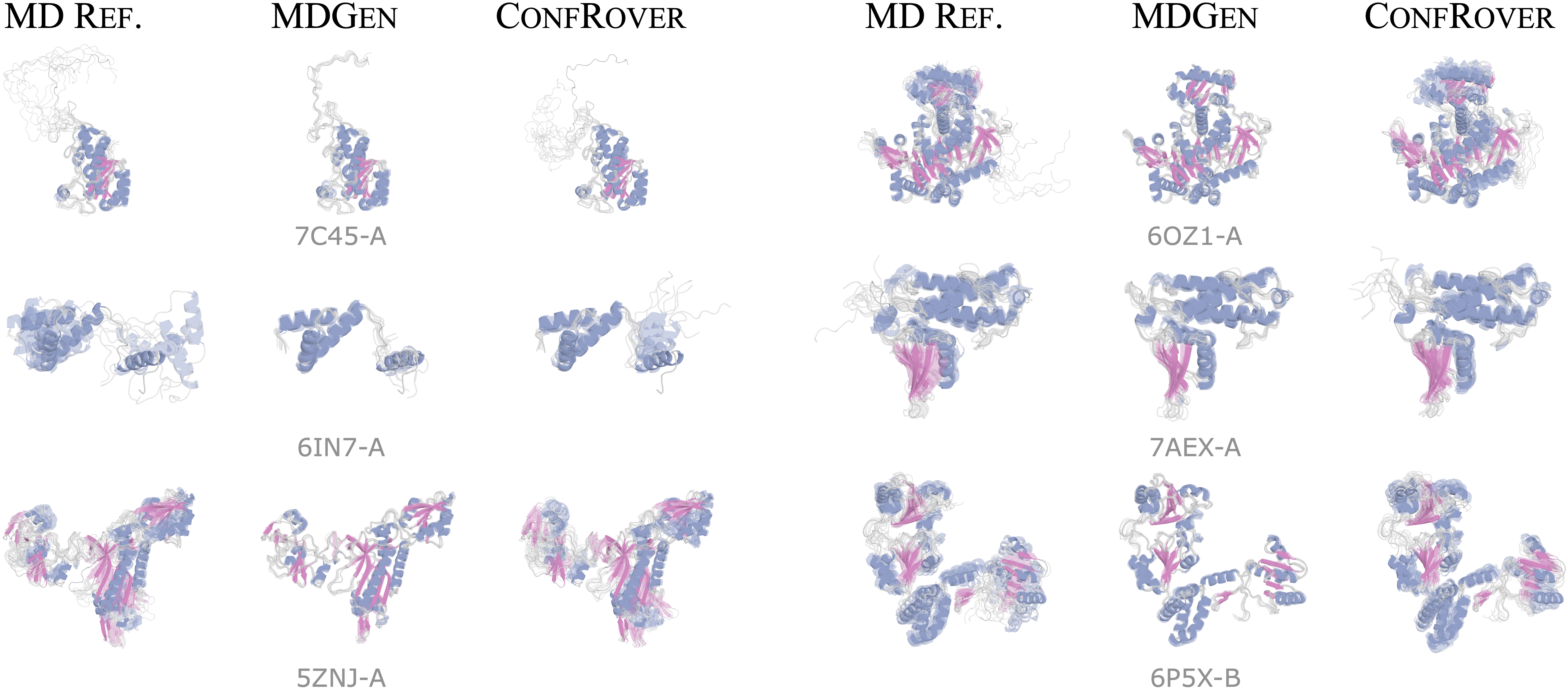}
    \caption{Visualization of six proteins from \textit{multi-start}. Trajectory conformations are colored by their secondary structures and superposed to show the dynamic ensemble. \textsc{MDGen} primarily exhibits local movements, whereas \model captures conformations changes similar to MD simulations.}
    \label{fig:multi-start-case}
    \vspace{-6pt}
\end{figure}

\textit{ConfRover shows superior performance in recovering the magnitude of conformational changes.} We report the Pearson correlation of measured conformation changes between model-generated and reference trajectories in Table~\ref{tab:fwd_vary_corr}, with additional results in Appendix~\ref{ap:multi-start}. Compared with \textsc{MDGen}, \model shows a significant improvement in correlation scores, mean absolute error and structural quality (Table~\ref{tab:fwd_vary_metrics}), indicating its stronger ability to recover the magnitude of conformation changes across different starting conditions in the conformational space. 
The greater difference observed in PCA highlights that \model more accurately captures conformational changes along the feature dimensions most relevant to the structural variance observed in MD.
Figure~\ref{fig:multi-start-case} visualizes ensembles of conformations in generated trajectories, with additional examples in Figure~\ref{fig:fwd_vary_cases_rnd}.
\model exhibit more notable conformation changes than \textsc{MDGen}, and reflects the major movements in the structured and loop domains observed in the MD reference. 

\textsc{MDGen} is a non-autoregressive model trained on trajectories of length $L=250$. Adjusting its inference setup results in degraded conformation. To ensure fair comparison, we use the original inference setting ($S = 40, L = 250$) and downsampled the trajectories for evaluation. To confirm that this post-processing step does not introduce artifacts, we also trained \textsc{MDGen} models under the evaluation setups. The results are consistent with the downsampled version (see Appendix~\ref{ap:mdgen_retrain}).

\begin{wraptable}{r}{0.44\textwidth}  
    \centering
    \vspace{-16pt}
    \footnotesize
    \captionof{table}{Recovery of conformational states in the 100~ns simulation experiment. The mean and standard deviation are computed over five independent runs, and the better results are highlighted in \textbf{bold}. \textsc{MD 100ns} serves as the oracle and is excluded from the comparison.}
    {
    \setlength{\tabcolsep}{3pt}        
    \renewcommand{\arraystretch}{1.1} 
    \begin{tabularx}{\linewidth}{cccc}
    \toprule
     & JSD ($\downarrow$) & Recall ($\uparrow$) & F1 ($\uparrow$) \\
    \midrule
    \rowcolor{Highlight}
    \textsc{MD 100ns} & 0.31 & 0.67 & 0.79 \\
    \textsc{MDGen} & \meanstd{0.56}{0.01} & \meanstd{0.29}{0.01}  & \meanstd{0.42}{0.01} \\
    \model & \textbf{\meanstd{0.51}{0.01}} & \textbf{\meanstd{0.42}{0.00}} & \textbf{\meanstd{0.58}{0.00}}\\
    \bottomrule
    \end{tabularx}
    \label{tab:fwd_100ns}
    }
    \vspace{-8pt}
\end{wraptable}

\noindent\textbf{Assessing long trajectory generation on 100~ns simulation.} We further evaluate model's ability to recover conformational states and principal dynamics of proteins. For each of protein, we simulate a trajectory of $L=80$ frames at stride $S=120$, approximating the 100~ns MD simulation in ATLAS. 
To assess state recovery, model-generated conformations are projected into a reduced PCA space and compared with the reference trajectory. Specifically, we discretize each principal component into 10 evenly sized ``states'' and measure the distribution similarity using Jensen-Shannon Distance (JSD). We also compute precision, recall, and F1-score on whether sample conformations fall within these known states~\citep{luStr2StrScorebasedFramework2024,wang2024proteinconfdiff,zhengPredictingEquilibriumDistributions2024aDiG}.
To evaluate dynamic mode recovery, we perform time-lagged independent component analysis (tICA) at varying lag times on both reference and sample trajectories. We then compute Pearson correlations between the per-residue contribution to the leading components, based on the tICA coefficients. See Appendix~\ref{ap:fwd_100ns} for details.
We include one of the triplicate MD trajectories—excluded from ground-truth evaluation—as an ``oracle'' reference, denoted as \textsc{MD 100ns}, representing the performance expected if the model were as accurate as an MD simulation run.

\begin{figure}[t]
    \centering
    \includegraphics[width=0.95\textwidth]{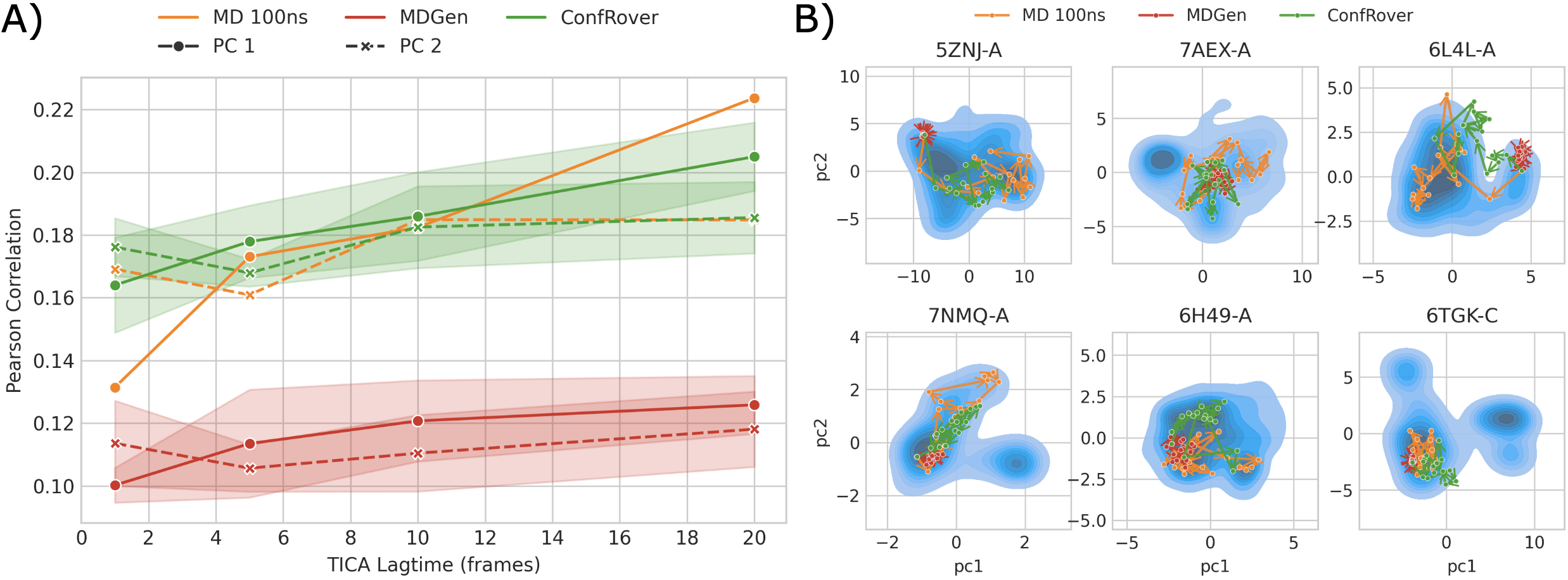}
    \caption{Results from 100~ns simulation. (A) Correlations of principal dynamic modes between sample and reference trajectories, evaluated at varying lag time. The mean and standard deviation are shown as line and shadowed area, computed from five individual runs for \textsc{MDGen} and \model. (B) Examples trajectories illustrating the states explored by different methods (downsampled by 5 frames for visualization). The blue background indicates the density of the ground-truth conformation distribution from MD reference.}
    \label{fig:fwd_100ns}
    \vspace{-8pt}
\end{figure}

\textit{ConfRover recovers more conformational states than MDGen and accurately captures the principal dynamics.} As shown in Table~\ref{tab:fwd_100ns}, \model outperforms \textsc{MDGen} in state recovery, achieving lower JSD, higher recall and F1 scores, showing its improved ability to capture diverse conformations. Additionally, \model shows clear advantage in capturing the principal dynamic modes across varying lag times, performing even comparably to the MD oracle (Figure~\ref{fig:fwd_100ns}A). This results suggest \model can learn and generalize dynamics to unseen proteins and still capture the most important dynamic modes.
We visualize simulated trajectories in the PCA space in Figure~\ref{fig:fwd_100ns}B and Figure~\ref{fig:fwd_100ns_rnd_case}. These examples again confirm that \model is more capable of sampling over the conformational space of the proteins and covering diverse conformations. Yet, we also observe some cases where \textsc{MD 100ns} overcame the energy barrier and achieved more remote states while \model did not (e.g., 7NMQ-A in Figure~\ref{fig:fwd_100ns}B).

\textit{Summary.} These experiments demonstrate that \model outperforms the current state-of-the-art model in trajectory simulation, effectively learning protein dynamics from MD data and generalizing well to unseen proteins.
While a gap remains compared to the oracle \textsc{MD 100ns}, particularly in state recovery, the improvement narrows the gap between deep generative models and established simulation methods.

\vspace{-4pt}
\subsection{Time-independent Conformation Sampling}
\vspace{-6pt}

We evaluate the time-independent sampling performance of \model, following the benchmark setup in~\citet{ye2024proteinbench}. For each protein, 250 independent conformations are sampled and compared the model generated ensembles with MD reference ensembles. We summarize the mean and standard deviations from five independent runs with key metrics are summarized in Table~\ref{tab:dist_main} with full results in Appendix~\ref{ap:dist}.

\noindent\textit{ConfRover matches the performance of state-of-the-art ensemble generation models.}
Compared with \textsc{AlphaFlow} and \textsc{ConfDiff}, \model demonstrate overall comparable performance and outperforms at least one of the SOTA models in five evaluation criteria. This demonstrate that \model, despite being a general-purpose model capable of trajectory generation, also performs strongly in sampling independent conformations that approximate the equilibrium distribution from MD simulation.
In contrast, \textsc{MDGen}, which is trained solely for trajectory generation, shows suboptimal results with sequentially sampled conformations.

\begin{table*}[h]
\centering
\vspace{-6pt}
\captionof{table}{Results from the time-independent generation experiment. \textsc{AlphaFlow} and \textsc{ConfDiff} are state-of-the-art models for direct conformation ensemble generation. \model is the base model trained for both trajectory and time-independent generation, where as \model[traj] and \textsc{MDGen} are trained exclusively for trajectory generation. The mean and standard deviation are computed from five independent runs. The best scores are highlighted in \textbf{bold}, and the second-best scores are \underline{underlined}.}
\label{tab:dist_main}
\setlength{\tabcolsep}{5pt}
{\fontsize{8pt}{9pt}\selectfont
\begin{tabular}{lcccccccc}
\toprule
& \specialcell{Pairwise\\RMSD\\$r$ ($\uparrow$)} & \specialcell{Per target\\RMSF\\$r$ ($\uparrow$)} & \specialcell{RMWD\\($\downarrow$)} & \specialcell{MD\\PCA\\$\mathcal{W}_2$ ($\downarrow$)}  & \specialcell{Joint\\PCA\\$\mathcal{W}_2$ ($\downarrow$)} & \specialcell{Weak\\contacts\\$J$ ($\uparrow$)} & \specialcell{Transient\\contacts\\$J$ ($\uparrow$)} & \specialcell{Exposed\\residue\\$J$ ($\uparrow$)} \\
\midrule
\textsc{AlphaFlow}  & \textbf{\meanstd{0.56}{0.06}} & \textbf{\meanstd{0.85}{0.01}} & \textbf{\meanstd{2.62}{0.03}} & \meanstd{1.52}{0.05}  & \meanstd{2.26}{0.03}  & \underline{\meanstd{0.62}{0.00}} & \textbf{\meanstd{0.41}{0.00}} & \textbf{\meanstd{0.69}{0.01}}\\
\textsc{ConfDiff}           & \underline{\meanstd{0.54}{0.00}} & \textbf{\meanstd{0.85}{0.00}} & \meanstd{2.70}{0.01} & \underline{\meanstd{1.44}{0.00}} & \textbf{\meanstd{2.22}{0.04}} & \textbf{\meanstd{0.64}{0.00}} & \underline{\meanstd{0.40}{0.00}} & 	\underline{\meanstd{0.67}{0.00}} \\
\midrule
\textsc{MDGen}             & \meanstd{0.47}{0.04} & \meanstd{0.72}{0.02} & \meanstd{2.78}{0.04} & \meanstd{1.86}{0.03} & \meanstd{2.44}{0.04} & \meanstd{0.51}{0.01} & \meanstd{0.28}{0.01} & \meanstd{0.57}{0.01} \\
\model[traj] & \meanstd{0.48}{0.00} & \meanstd{0.84}{0.01} & \meanstd{2.85}{0.02} & \textbf{\meanstd{1.43}{0.01}} & \meanstd{2.30}{0.01} & \meanstd{0.53}{0.01} & \meanstd{0.36}{0.00} & \meanstd{0.58}{0.01} \\

\model   & \meanstd{0.51}{0.01} & \textbf{\meanstd{0.85}{0.00}} & \underline{\meanstd{2.66}{0.02}} & {\meanstd{1.47}{0.03}} & \underline{\meanstd{2.23}{0.04}} & \underline{\meanstd{0.62}{0.01}} & \meanstd{0.37}{0.01} & {\meanstd{0.66}{0.01}} \\
\bottomrule
\end{tabular}
}
\vspace{-6pt}
\end{table*}

\noindent\textbf{Effect of hybrid training.} Without explicit single-frame training, the model primarily learns time-dependent generation, with only the first frame of each trajectory learning to generate conformation unconditionally (i.e., from a masked token input). To test the importance of hybrid training, we ablated the single-frame objective and trained a variant, \model[traj], solely on trajectory generation. As shown in Table~\ref{tab:dist_main}, while this variant still outperforms time-dependent results from \textsc{MDGen}, it shows decreased performance across several metrics compared to \model. This highlights the importance of hybrid training in balancing the learning objectives and enhancing the model's for generating independent conformations.

\subsection{Conformation Interpolation}

To enable \model for conformation interpolation, we continue training \model with a hybrid objective combining trajectory, single-frame and interpolation, referred as \model[interp]. We select 38 short trajectories from \textit{multi-start} for evaluation where the reference MD trajectories for these cases exhibit significant conformation changes and clear state transitions, see Appendix~\ref{ap:interp} for details. To condition on both start and end frames, we prepend the end frame to the start frame and autoregressively generate the remaining (intermediate) frames. To evaluate whether the model generate smooth transitions towards the target end state, we measure C$\alpha$-RMSD and $L^2$~distance in the PCA space between each intermediate frame and the start/end frames.

\begin{figure}[hb]
    \vspace{-12pt}  
    \centering
    \includegraphics[width=1.0\linewidth]{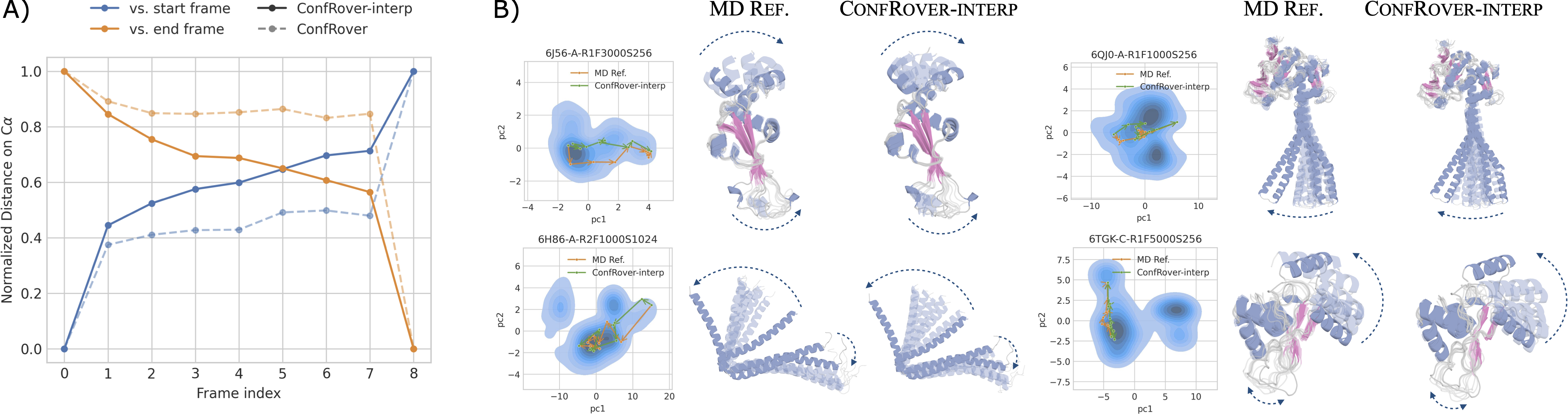}
    \caption{Results from \textit{conformation interpolation.} (A) C$\alpha$-RMSD distance of intermediate frames to the start and end frames, normalized by the distance between start and end frames. Reported values are averaged over 38 cases selected from the \textit{multi-start} benchmark. (B) Example interpolations results. \model[interp] generates smooth pathways between the start and end frames, capturing the dynamics observed with the MD reference. Start and end frames are shown as solid structures; intermediate conformations are shown in fading colors. Main motions are indicated by blue arrows.}
    \label{fig:interp_main}
    \vspace{-6pt}  
\end{figure}

\textit{Training on the interpolation objective enables smooth interpolation between conformations.} As shown in Figure~\ref{fig:interp_main}A, the distance to the start frame increases while the distance to the end frame decreases with frame index, indicating smooth and directed transitions. Without explicit interpolation training, the original \model (dashed lines in Figure~\ref{fig:interp_main}A) generates trajectory that do not progress towards the end state.
Figure~\ref{fig:interp_main}B visualizes intermediate structures and transition pathways in PCA space, showing that intermediate conformations from \model[interp] closely resemble those in the MD reference. In contrast, as shown in Figure~\ref{fig:interp_add_vis}, the original \model can miss key transitions and fails to reach the end state. Additional results and visualizations are provided in Appendix~\ref{ap:interp}. These results highlight the effectiveness of our interpolation training strategy: by adjusting the dependency order in the sequence model, \model[interp] learns to generate smooth transitions between two conformations.

\setlength{\abovecaptionskip}{2pt}

\subsection{Conformation Quality}
\vspace{-6pt}

To ensure that \model generates physically plausible conformations, we further evaluate the quality using geometric assessments from MolProbity package~\citep{williams2018molprobity} and energy profiles using a coarse-grained force field MadraX~\citep{orlando2024madrax}. 
We compared conformations across 38 trajectories shared between the forward simulation and interpolation experiments, including results from \textsc{MDGen}, \model, and an oracle \textsc{MD reference}. All structures are relaxed using the refinement pipeline in OpenFold~\citep{ahdritzOpenFoldRetrainingAlphaFold22023} to enable energy comparison.

As shown in Table~\ref{tab:quality}, conformations generated by MD simulation exhibit the hightest overall quality, as expected.
As a generative model, \model also produces high-quality conformations with fewer backbone (Ramachandran) and side-chain (rotamer) outliers, more accurate covalent lengths and angles, and achieving energy levels comparable to those of \textsc{MD Reference}; outperforming \textsc{MDGen} across all metrics.
Furthermore, we compare conformations generated from forward simulation and interpolation tasks, where the latter includes additional constraints on terminal conformations. \model[interp] shows similar geometric and energetic metrics, indicating that intermediate conformations maintains physical plausible when \model tries to interpolate between two conformational states.

\begin{table*}[t]
\centering
\vspace{-6pt}
\captionof{table}{Quality of model generated conformations. Conformations sampled from 38 trajectories are evaluated. Geometric metrics are reported as mean and standard deviations across 38 cases, and energy values are reported as mean with 95\% percentiles. The best scores are highlighted in \textbf{bold}.}
\label{tab:quality}
\setlength{\tabcolsep}{2pt}
{\fontsize{8pt}{9pt}\selectfont
\begin{tabular}{@{}lcccccccc@{}}
\toprule
& 
\specialcell{Ramachandran\\outliers \% ($\downarrow$)} & 
\specialcell{Rotamer\\outliers \% ($\downarrow$)} & 
\specialcell{Clash\\score ($\downarrow$)} & 
\specialcell{RMS\\bonds ($\downarrow$)} & 
\specialcell{RMS\\angles ($\downarrow$)} & 
\specialcell{MolProbity\\score ($\downarrow$)} & 
\specialcell{MadraX\\energy ($\downarrow$)} \\
\midrule
\rowcolor{Highlight}
\textsc{MD Reference}    & \meanstd{0.38}{0.49} & \meanstd{1.02}{0.89} & \meanstd{0.04}{0.16} & \meanstd{0.01}{0.00} & \meanstd{1.88}{0.05} & \meanstd{0.72}{0.18} & -519.3 {\tiny (-1793.0,-53.4)} \\
\textsc{MDGen}      & \meanstd{0.93}{0.86} & \meanstd{2.86}{1.59} & \meanstd{16.14}{20.05} & \meanstd{0.02}{0.02} & \meanstd{2.13}{0.30} & \meanstd{2.24}{0.40} & -314.7 {\tiny (-1483.8,263.6)} \\
\model              & \textbf{\meanstd{0.58}{0.63}} & \meanstd{1.98}{1.48} & \meanstd{7.81}{6.52} & \textbf{\meanstd{0.01}{0.01}} & \textbf{\meanstd{1.88}{0.25}} & \meanstd{1.72}{0.38} & \textbf{-522.2 {\tiny (-1858.9, -53.4)}} \\
\model[interp]      & \meanstd{0.71}{0.94} & \textbf{\meanstd{1.86}{1.46}} & \textbf{\meanstd{7.25}{8.74}} & \meanstd{0.02}{0.01} & \meanstd{1.91}{0.32} & \textbf{\meanstd{1.61}{0.51}} & -469.7 {\tiny (-1712.3, -42.8)} \\
\bottomrule
\end{tabular}
}
\vspace{-6pt}
\end{table*}

\vspace{-4pt}
\section{Conclusions and Limitations}
\vspace{-4pt}
\label{sec:limit}
We introduce \model, a general framework for learning protein conformational dynamics from MD trajectory data. Through autoregressive factorization, \model supports three tasks in a unified manner: trajectory simulation, time-independent sampling, and conformation interpolation. This formulation reflects the temporal nature of MD while naturally encompassing conditional and unconditional frame-level generation. Extensive experiments and analyses highlight several empirical advantages: (1) \model outperforms the current state-of-the-art in trajectory simulation, accurately capturing dynamic magnitude, state recovery, and principal motions; (2) Despite being a multi-purpose model, it achieves competitive performance in time-independent sampling compared to specialized methods; (3) With simple sequence reordering, \model effectively learns to interpolate between conformations.

Nevertheless, several limitations still remain: 
(1) Trajectory simulation and interpolation are emerging tasks with few available baseline models. We hope that this work, together with future developments in the field, will contribute to establishing more comprehensive benchmarks;
(2) The dataset and evaluation metrics used in this study are limited and preliminary. The ATLAS dataset contains 100~ns simulations of single-chain proteins, which may not capture large conformational changes or the dynamics of protein complexes. Although we curated interpolation cases from ATLAS for demonstration purposes, future benchmarks reflecting realistic functional state transitions would be more meaningful.
(3) Although \model narrows the gap with classical MD, it still falls short in fully capturing the conformational space with high structural fidelity. Future gains may come from scaling training data, using more efficient architectures, and leveraging additional information from MD, such as energy information.
(4) Finally, while the triangular updates in the structural modules ensure high accuracy, their computational cost limits scalability to larger proteins and longer trajectories.
Despite these challenges, \model demonstrates the promise of autoregressive models in molecular simulation, offering a unified, efficient, and extensible approach to modeling protein dynamics.

\section{Acknowledgments}
\vspace{-6pt}
We would like to thank Dr. Hang Li for his invaluable support of this project. We also thank Zaixiang Zheng for insightful discussions, and Wesley Hsieh, Yi Zhou, Nima Shoghi, Yuxuan Liu, Xiaolu Shen, Jing Yuan, Yilai Li, Fei Ye, and Wei Qu for their valuable feedback.

\clearpage
\newpage

\bibliographystyle{plainnat}
\bibliography{reference}

\clearpage
\newpage
\appendix

\section{Additional Background}
\label{ap:discuss}

\subsection{Related Work}
\paragraph{Deep Generative Models for MD Trajectories.}
Recent works have explored generating protein trajectories as a surrogate for MD simulations. Models such as \textsc{Timewarp}~\citep{klein2024timewarp}, \textsc{ITO}~\citep{schreiner2024implicit}, and \textsc{EquiJump}~\citep{costa2024equijump} learn stochastic transport functions to sample future conformations at a lagged time (longer than MD intervals), reducing the computational costs of long-timescale simulations. However, these methods assume Markovian dynamics by relying solely on the current state for prediction, which may not be suitable for non-Markovian dynamics common in protein MD data.
To capture higher-order dependencies between the frames, \textsc{AlphaFolding}~\citep{cheng20244d4ddiff} incorporates history frames via ``motion nodes'', but it requires a fixed context window. 
\textsc{MDGen}~\citep{jing2024generative_mdgen} takes a different approach by directly modeling the joint distributions of frames in a trajectory and learning frame dependencies through ``masked frame modeling'', similar to masked language modeling. However, its key-frame parameterization requires separate models for different tasks, and its non-autoregressive paradigm limits flexible generation (e.g., not compatible for generating trajectories with variable lengths).
\textsc{GST}~\citep{li2025geometricGST} applies autoregression for future frame prediction, enabling variable-length conditioning context and prediction horizons.
While the autoregressive approach is conceptually similar to \model, their work differs in several key aspects: it performs deterministic prediction rather than generative trajectory sampling; it employs a graph-based architecture with fixed structural priors from an adjacency graph, instead of full attention across all residues; it is trained and evaluated on a single protein instead of diverse proteins from ATLAS under a transferable setting.

Beyond forward trajectory simulation, generative models have also been applied to conformation interpolation, that is, generating the intermediate trajectories between two conditioned states. The non-autoregressive framework of \textsc{MDGen}~\citep{jing2024generative_mdgen} can be extended to sample transition pathways between such states; however, its key-frame parameterization requires training a separate model for this task, and it has not been tested on large proteins like those in ATLAS. 
\citet{du2024doobs} proposed a simulation-free objective for transition-pathway sampling based on Doob’s $h$-transform, but their approach has only been validated on numerical models and the small protein Chignolin. Its generalizability to larger, more diverse proteins remains unassessed.

Notably, the above models focus on learning temporal dependencies between frames and do not support direct, time-independent conformation sampling from the learned distribution. In contrast, \model is a general framework that learns both the trajectory generation tasks as well as direct sampling of independent protein conformations.

\paragraph{Deep Learning Models for Conformation Ensemble Generation.}
Another line of work focuses on direct sampling of conformations in a time-independent manner.
Early efforts include perturbing the input to folding models (e.g., AlphaFold)~\citep{delalamoSamplingAlternativeConformational2022MSAsampling,wayment-steelePredictingMultipleConformations2024MSA,steinSPEACH_AFSamplingProtein2022MSA} or perturbing the conformation using a structural diffusion model~\citep{luStr2StrScorebasedFramework2024}. 
However, these models are trained solely on static PDB structures and do not explicitly model the conformational distribution.
Recent works have shifted to deep generative paradigms that directly learn protein-specific conformational distributions~\citep{jingEigenFoldGenerativeProtein2023EigenFold,zhengPredictingEquilibriumDistributions2024aDiG,jingAlphaFoldMeetsFlow2024,wang2024proteinconfdiff,lu2024structureesmdiff,lewis2024scalableBioemu}. 
Several models in this category, including \textsc{AlphaFlow}, \textsc{ConfDiff}, and \textsc{BioEmulator}, fine-tune pretrained structure models on large-scale MD datasets, enhancing their ability to capture conformational distributions. A related approach trains normalizing flow models to approximate the Boltzmann distribution~\citep{noe2019boltzmann,kohler2020equivariant}, but their invertibility constraints limit scalability and transferability beyond small molecules and peptides. 
While these methods can generate time-independent conformations, they overlook temporal relationships and do not capture the kinetic aspects of protein dynamics.

\paragraph{Deep Learning Enhanced Molecular Dynamics.} Another direction integrates deep learning with molecular conformation modeling through machine learning force field (MLFF) models~\citep{kovacs2025maceoff,rhodes2025orbv3atomisticsimulationscale}. These models aim to incorporate higher-level accuracy (e.g., from \textit{ab initio} calculations) into classical molecular dynamics simulations, improving fidelity while maintaining scalability. 
However, they still rely on sequential MD sampling with small integration steps and can be more computationally expensive than conventional MD.
Two-for-One~\citep{Arts2023Twoforone} does not directly learn an MLFF but instead trains a diffusion model for protein conformations and uses the resulting score function as an approximate coarse-grained force field for MD simulation. Although it still depends on sequential MD sampling, this work provides an interesting perspective that connects conformation distribution modeling and trajectory generation. 
In contrast, different from these approaches, \model explicitly models and generates both individual conformation and trajectories, without relying on force field-based MD simulations.

\paragraph{Image-Video Generation.}
The challenge of modeling protein dynamics conceptually parallels tasks in image and video generation, requiring both data distribution learning and temporal modeling. Recent advances in video generation offer valuable insights in addressing these challenges. 
Given limited video data, extending image generative models to video has proven effective. Several works~\citep{ho_video_2022,blattmann2023stablevideo,ho2022imagen} achieve this by incorporating temporal attention layers, enabling frame-to-frame communication. Disabling temporal attention reverts them to image models, allowing flexible training across both modalities. These approaches efficiently model time correlations without explicitly tracking offsets between frames. 
Meanwhile, the extension of autoregressive language models to image and video domains has shown strong potential for sequential generation in different data modalities. \citet{li2024autoregressivemar} integrates language models’ sequential modeling with diffusion models’ ability to model continuous distributions, showing that discrete tokens are not essential for autoregressive models. \textsc{MarDini}~\citep{liu2024mardini} extends the concept to video generation with efficient llama-style temporal planning and high-resolution video generation via a diffusion decoder. By applying masked ``frame'' modeling, \textsc{MarDini} allows the model to learn flexible temporal relationships and enables diverse tasks such as frame interpolation.
\model differs from these works from video models in several aspects: it employs SE(3) diffusion for 3D structure generation; by using an autoregressive paradigm, it explicitly decouples the diffusion generation from temporal modeling, unlike the spatiotemporal denoising process in \textsc{MarDini}; in addition, the causal autoregressive framework enables more flexible trajectory generation with variable lengths.

\subsection{Baseline Limitation}

Modeling MD trajectories using deep generative models is still an emerging research area, few models currently support learning protein dynamics in transferable settings. Existing approaches based on forward transport operators have been primarily trained and evaluated on small peptides (e.g., Timewarp~\citep{klein2024timewarp}) or small fast-folding proteins (e.g., EquiJump~\citep{costa2024equijump}). Although AlphaFolding~\citep{cheng20244d4ddiff} was also trained using ATLAS, the model weights are not publicly available at the time of this work. (We attempted an internal reproduction and included its result in Appendix~\ref{ap:alphafolding}.) Due to these limitations, we use \textsc{MDGen} as the only available model for the main experiment. 
For conformation interpolation, neither sampling-based method~\citep{du2024doobs} nor video-like method \textsc{MDGen}~\citep{jing2024generative_mdgen} have been trained and evaluated on large proteins. Therefore, we focus on demonstrating the interpolation results of \model[interp].

\clearpage
\newpage
\section{Diffusion Models on SE(3) Space}
\label{ap:se3_diff}

Diffusion Probabilistic Models (DPM) model complex distributions through iterative denoising. In the context of protein conformations, DPMs defined over $\mathrm{SE(3)}$ translation-rotation space have been applied for protein backbone structural generation \citep{yimSEDiffusionModel2023,wang2024proteinconfdiff}. 
Following Section \ref{sec:prot_rep}, $\mathbf{x}_0 = (\mathbf{T}_0, \mathbf{R}_0) \in \mathrm{SE(3)}^N$ denotes the translations and rotations of backbone \textit{rigids} in data. The diffusion processes defined in the translation and rotation subspace add noise to corrupt the data:

\begin{align}
\mathrm{d} \mathbf{T}_t & = -\frac{1}{2} \beta_t \mathrm{P}\mathbf{T}_t \mathrm{~d} t+\sqrt{\beta_t} \mathrm{P} \mathrm{d} \mathbf{w}_t, \nonumber \\
\mathrm{~d}\mathbf{R}_t & = \sqrt{\frac{\mathrm{~d}}{\mathrm{~d}t} \sigma^2_t}\mathrm{~d} \mathbf{w}_t^{\mathrm{SO(3)}} \nonumber,
\end{align}
where $t \in [0, 1]$ is the diffusion time, $\beta_t$ and $\sigma_t$ are predefined time-dependent noise schedules and $\mathrm{P}$ is a projection operator removing the center of mass. $\mathbf{w}_t$ and $\mathbf{w}_t^{\mathrm{SO(3)}}$ are the standard Wiener processes in $\mathcal{N}(0, I_3)^{\otimes N}$ and $\mathcal{U}(\mathrm{SO}(3))^{\otimes N}$ respectively.

The transition kernel of $\mathbf{T}$ satisfies $p_{t|0}(\mathbf{T}_t|\mathbf{T}_0)=\mathcal{N}(\mathbf{T}_t; \sqrt{\alpha_t} \mathbf{T}_0,(1-\alpha_t)\mathbf{I})$, where $\alpha_t = e^{-\int_0^t\beta_sds}$. The transition kernel of $\mathbf{R}$ satisfies $p_{t|0}(\mathbf{R}_t|\mathbf{R}_0)=\mathcal{IGSO}_3(\mathbf{R}_t;\mathbf{R}_0, t)$,
where $\mathcal{IGSO}_3$ is the isotropic Gaussian distribution on $\mathrm{SO(3)}$ \citep{yimSEDiffusionModel2023}.

The associated reverse-time stochastic differential equation (SDE) follows:
\begin{align}
\mathrm{d} \mathbf{T}_t & = \mathbf{P}\left[-\frac{1}{2} \beta_t \mathbf{T}_t - \beta_t \nabla\log p_t(\mathbf{T}_t)\right]\mathrm{d}t +\sqrt{\beta_t}\mathrm{P} \mathrm{d}\bar{\mathbf{w}}_t, \nonumber \\
\mathrm{d} \mathbf{R}_t & = - \frac{\mathrm{~d}}{\mathrm{~d}t} \sigma^2_t \nabla\log p_t(\mathbf{R}_t) \mathrm{d} t+\sqrt{\frac{\mathrm{d}}{\mathrm{d}t} \sigma^2_t}\mathrm{d} \bar{\mathbf{w}}_t^{\mathrm{SO(3)}},
\label{eq:rev_diff}
\end{align}

where $\bar{\mathbf{w}}_t$ and $\bar{\mathbf{w}}_t^{\mathrm{SO(3)}}$ denote standard Wiener processes in the reverse time.

The reverse process can be approximated by a neural network through the \textit{denoising score matching} loss for translation and rotation:
\begin{align}
\mathcal{L}(\theta) =& \mathcal{L}^{\mathbf{T}}(\theta) + \mathcal{L}^{\mathbf{R}}(\theta) \label{eq:dsm} \\ 
=& \mathbb{E}\left[ \lambda(t) \| s_\theta(\mathbf{T}_t, t) - \nabla_{\mathbf{T}_t} \log p_{t|0}(\mathbf{T}_t | \mathbf{T}_0) \|^2\right] \nonumber \\ 
&+ \mathbb{E}\left[
\lambda^r(t) \| s_\theta^r(\mathbf{R}_t, t) - \nabla_{\mathbf{R}_t} \log p_{t|0}(\mathbf{R}_t | \mathbf{R}_0) \|^2 \right] \nonumber ,
\end{align}
where $\lambda(t)$ and $\lambda^r(t)$ are time-dependent weights, $s_\theta(\mathbf{T_t, t})$ and $s^r_\theta(\mathbf{R}_t, t)$ are the score networks commonly parameterized with shared weights. The expectations are taken over diffusion time $t\sim \mathcal{U}[t_\mathrm{min}, 1]$, and over noisy and clean data pairs from the forward process $(\mathbf{T}_0, \mathbf{T}_t)$ and $(\mathbf{R}_0, \mathbf{R}_t)$.

\clearpage
\section{Derivation of Equations in Section~\ref{sec:latent_causal_modeling}}
\label{ap:latent_factorize}

Here we provide a more rigorous derivation of equations in 
 Section~\ref{sec:latent_causal_modeling}. For clarity, we omit the conditioning variable $\mathcal{P}$ in the intermediate steps.

As defined in Equation~\eqref{eq:traj_decomp}, our goal is to model the frame-level conditional distribution $p(\mathbf{x}^l|\mathbf{x}^{<l})$. To achieve this, we encode the previous frames $\{\mathbf{x}^i\}_{i=1}^{l-1}$ into a sequence of latent embeddings $\{\mathbf{h}^i\}_{i=1}^{l-1}$, and model inter-frame dependencies in this latent space.

By applying the Bayes' rule, we can factor the joint distribution over the current conformation $\mathbf{x}^{l}$ and intermediate latent embeddings $\mathbf{h}=(\mathbf{h}^{l}, \mathbf{h}^{<l})$ as:
\begin{equation*}
    p(\mathbf{x}^l, \mathbf{h}^l, \mathbf{h}^{<l} \mid \mathbf{x}^{<l}) = p(\mathbf{x}^l \mid \mathbf{h}^l, \mathbf{h}^{<l}, \mathbf{x}^{<l}) \, p(\mathbf{h}^l \mid \mathbf{h}^{<l}, \mathbf{x}^{<l}) \, p(\mathbf{h}^{<l} \mid \mathbf{x}^{<l}).
\end{equation*}

Integrating both sides over the latent variables $\mathbf{h}$ yields:
\begin{equation}
\label{eq:joint_margin_full}
    p(\mathbf{x}^l \mid \mathbf{x}^{<l}) = \int_{\mathbf{h}} p(\mathbf{x}^l \mid \mathbf{h}^l, \mathbf{h}^{<l}, \mathbf{x}^{<l}) \, p(\mathbf{h}^l \mid \mathbf{h}^{<l}, \mathbf{x}^{<l}) \, p(\mathbf{h}^{<l} \mid \mathbf{x}^{<l}) \, \mathrm d\mathbf{h}.
\end{equation}

In our approach, both $p(\mathbf{h}^l \mid \mathbf{h}^{<l}, \mathbf{x}^{<l})$ and $p(\mathbf{h}^{<l} \mid \mathbf{x}^{<l})$ are modeled using deterministic neural networks: an encoder $f_\eta^{\text{enc}}(\mathbf{x}^i)$ and an autoregressive temporal module $f_\xi^{\text{temp}}(\mathbf{h}^{<l})$, respectively.

These mappings reduce both $p(\mathbf{h}^l \mid \mathbf{h}^{<l}, \mathbf{x}^{<l})$ and $p(\mathbf{h}^{<l} \mid \mathbf{x}^{<l})$ to Dirac delta functions, and the conditional dependencies can be simplified as:
\begin{align*}
    &p(\mathbf{h}^l |\mathbf{h}^{<l}, \mathbf{x}^{<l}) = p(\mathbf{h}^l |\mathbf{h}^{<l}) \\
    &p(\mathbf{x}^l |\mathbf{h}^l , \mathbf{h}^{<l}, \mathbf{x}^{<l}) = p(\mathbf{x}^l |\mathbf{h}^l ).
\end{align*}

Substituting into Equation~\eqref{eq:joint_margin_full} gives:
\begin{align}
\label{eq:joint_margin}
    p(\mathbf{x}^l | \mathbf{x}^{<l})
        &=
    \int_{\mathbf{h}}p(\mathbf{x}^l | \mathbf{h}^l)
    \cdot 
    p(\mathbf{h}^l | \mathbf{h}^{<l})
    \cdot 
    p(\mathbf{h}^{<l} | \mathbf{x}^{<l})\mathrm{d}\mathbf{h}.
\end{align}

Again, due to the deterministic nature of the encoder and temporal module, there is no marginalization involved in Equation \eqref{eq:joint_margin}, yielding:

\begin{align*}
p(\mathbf{x}^l|\mathbf{x}^{<l}) &= p(\mathbf{x}^{l}|\mathbf{h}^{l}) \\
\text{where} \quad \mathbf{h}^i &= f_\eta^\text{enc}(\mathbf{x}^i, \mathcal{P}), \quad i=1, 2, \dots, l-1 \\
\text{and} \quad \mathbf{h}^{l} &= f_\xi^\text{temp} \left(\mathbf{h}^1, \mathbf{h}^2, \dots, \mathbf{h}^{l-1}\right).
\end{align*}

Finally, we approximate $p(\mathbf{x}^{l}|\mathbf{h}^{l})$ with a parameterized model $p_\theta^{\text{enc}}(\mathbf{x}^{l}|\mathbf{h}^{l})$.

\clearpage

\section{Method Details}
\label{ap:mtd_details}

\subsection{Detailed Module Architectures}
\label{ap:arch_details}

\paragraph{Encoding Layer.}

The protein-specific single and pair representations are obtained from the Evoformer stack of a pretrained OpenFold model (with frozen weights), after three recycle iterations. In addition, we encode residue-level sequence information by combining sinusoidal positional embeddings of residue indices with learnable embeddings for the 20 standard amino acid types. These features are concatenated with the single representation from the \texttt{FoldingModule}.

To encode the structural information of each conformation frame, we introduced the \texttt{FrameEncoder}, a pseudo-beta-carbon (C$_\beta$)  coordinate encoder similar to the \texttt{InputEmbedding} module from \textsc{AlphaFlow}~\citep{jingAlphaFoldMeetsFlow2024} (without diffusion time embedding). Specifically, this module first compute the pairwise distances between residues using C$_\beta$ coordinates. These distances are then binned, embedded into latent embedding, and further refined through triangular update blocks including triangle attention and multiplication updates \citep{jumperHighlyAccurateProtein2021AF2}. See Algorithm \ref{alg:embedding} for the specifics. The resulting per-frame C$_\beta$ pair embedding $\mathbf{z}_\text{Frame}^l$ is concatenated with the pair representation from \texttt{FoldingModule}.

\begin{figure*}[h]
    \centering
    \includegraphics[width=\linewidth]{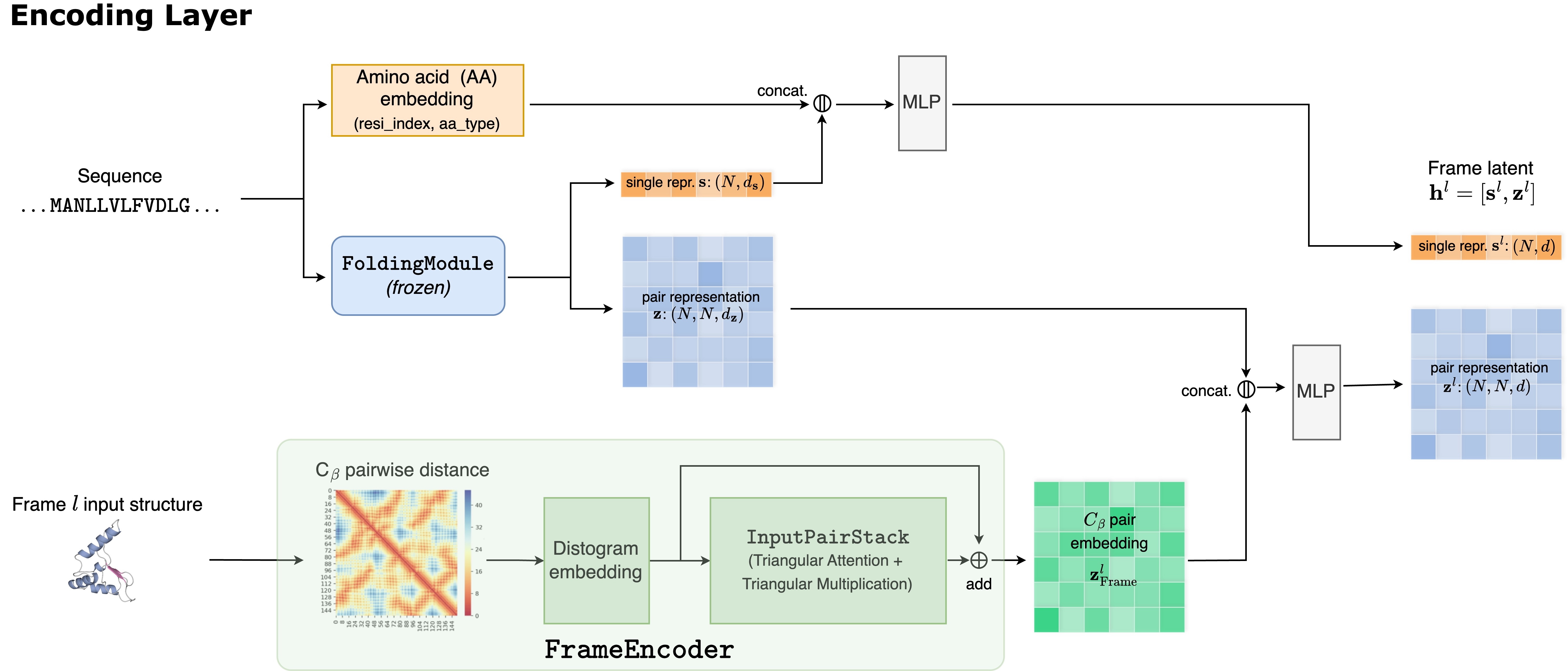}
    \caption{Architecture details of the \textit{Encoding Layer}. A frozen \texttt{FoldingModule} encodes the protein-specific information from its sequence, containing prior knowledge on its chemical environment and folding structures. The single representation is further concatenated with additional amino acid embeddings and projected to a hidden dimension of size $d$; The pair representation is concatenated with frame conformation information, encoded in C$_\beta$ pair embedding, and projected to a hidden dimension of size $d$. Both frame-level single and pair embeddings form the frame-level latent for downstream modules.}
    \label{fig:arch_enc}
\end{figure*}

Both single and pair embeddings are projected into the same dimension of $d$ for simplicity, forming the latent embedding $\mathbf{h}^l=[\mathbf{s}^l, \mathbf{z}^l]$ for each frame. See detailed illustration in Figure \ref{fig:arch_enc}.

\begin{algorithm}[H]
\caption{\textsc{FrameEncoder}}
\label{alg:embedding}
\begin{algorithmic}
\STATE \textbf{Input:} {Pseudo beta carbon (C$_\beta$) coordinates $\bfx \in \mathbb{R}^{N \times 3}$, time $t\in[0,1]$}
\STATE \textbf{Output:} {Input pair embedding $\bfz \in \mathbb{R}^{N \times N \times 64}$}
\STATE $\bfz_{ij} \gets \lVert \bfx_i - \bfx_j \rVert $ 
\STATE $\bfz_{ij} \gets \Bin(\bfz_{ij}, {\min}=3.25 \text{ \AA}, {\max}= 50.75 \text{ \AA},  N_\text{bins}=39)$ 
\STATE $\bfz_{ij} \gets \Linear(\OneHot(\bfz_{ij}))$ 
\FOR{$l \gets 1$ to $N_\text{blocks} = 4$}
\STATE    $\{\bfz\}_{ij} \pluseq \TriangleAttentionStartingNode({\bfz_{ij}}, c = 64, N_\text{head} = 4)$   
\STATE    $\{\bfz\}_{ij} \pluseq \TriangleAttentionEndingNode({\bfz_{ij}}, c = 64, N_\text{head} = 4))$ 
\STATE    $\{\bfz\}_{ij} \pluseq \TriangleMultiplicationOutgoing({\bfz_{ij}}, c = 64)$   
\STATE    $\{\bfz\}_{ij} \pluseq \TriangleMultiplicationIncoming({\bfz_{ij}}, c = 64)$ 
\STATE    $\{\bfz\}_{ij} \pluseq \PairTransition({\bfz_{ij}}, n = 2)$ 
\ENDFOR
\STATE $\bfz_{ij} = \LayerNorm(\bfz_{ij})$

\end{algorithmic}
\end{algorithm}

\paragraph{Trajectory Module.}
In the \textit{Trajectory Module}, we interleave layers of \texttt{StructuralUpdate} and \texttt{TemporalUpdate} to iteratively update the latent $[\mathbf{s}^l, \mathbf{z}^l]$, enabling the temporal reasoning across frames and structural refinement within each frame.

For the \texttt{StructuralUpdate}, we adopt a \texttt{Pairformer} block from AlphaFold 3~\citep{ahdritzOpenFoldRetrainingAlphaFold22023}, which  jointly updates the single and pair embeddings of the current frame through structural reasoning. After the \texttt{StructuralUpdate}, the pair embedding is flattened from $[N, N, d]$ to $[N\times N, d]$ and concatenated with the single embedding before being passed into the \texttt{TemporalUpdate}.

To model temporal dependencies between frames, we use a lightweight Llama architecture~\citep{touvron2023llama}. We transpose the input such that the temporal dimension is treated as the sequence axis for channel-wise self-attention across time. Rotary positional encoding~\citep{su2023roformerenhancedtransformerrotary} is applied to encode the temporal position for each frame. A causal attention mask is applied to restrict each frame to only attend to previous frames. After the temporal update, the latent embedding are reshaped and split back into single and pair embeddings.

\begin{figure*}[h]
    \vspace{12pt}
    \centering
    \includegraphics[width=\linewidth]{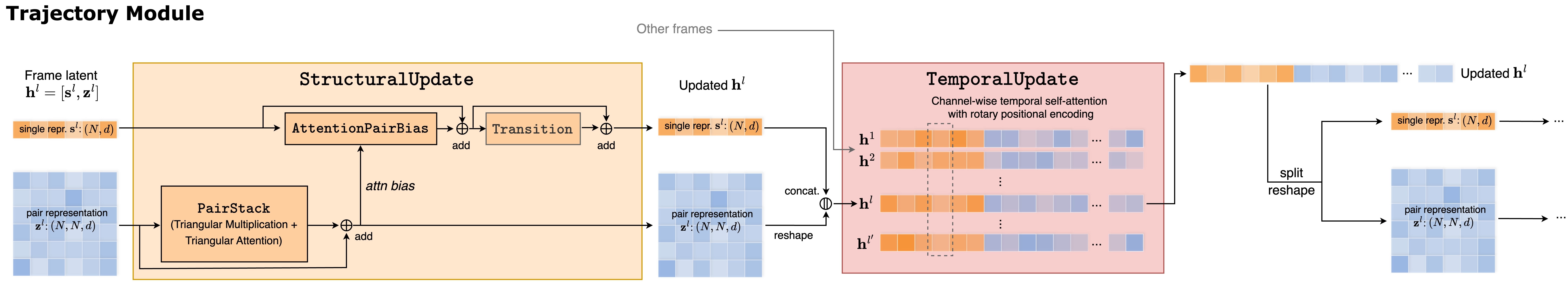}
    \caption{Architecture details of the \textit{Trajectory Module}. \textit{Trajectory Module} contains interleaving blocks of \texttt{StructuralUpdate} and \texttt{TemporalUpdate} (only one block of each is shown). \texttt{StructuralUpdate} leverages the \textit{Pairformer} architecture from~\citet{abramsonAccurateStructurePrediction2024}, updating the pair embeddings with triangular updates and the single embeddings using with pair bias from the updated pair. The updated pair embeddings are flatten and concatenate with single embedding for channel-wise temporal update. The attention is applied along the temporal dimension and update each single and pair embedding channels independently. The embeddings from \texttt{TemporalUpdate} are split and reshape back into single and pair embeddings.}
    \label{fig:arch_traj}
    \vspace{12pt}
\end{figure*}

Figure \ref{fig:arch_traj} and Table \ref{tab:hyperparameters_traj} provide the detailed module architecture and hyperparameter configurations, respectively. A \texttt{StructuralUpdate} block is included for every two \texttt{TemporalUpdate} layers.

\begin{table*}[hbt!]
 \label{tab:hparam1}
\centering
\caption{Hyperparameter choices of \textit{Trajectory Module}}
\setlength{\tabcolsep}{16mm
}
{\fontsize{8pt}{9pt}\selectfont
\begin{tabular}{ll}
\toprule[1.2pt]
\textbf{Hyperparameters} & \textbf{Values}\\ 
 \midrule[0.5pt]
 \multicolumn{2}{c}{\textbf{Lightweight Llama} (\texttt{TemporalUpdate})} \\
 \midrule[0.5pt]
Number of layers & 8 \\
Dimension of the MLP embeddings & 256 \\
Dimension of the hidden embeddings & 128 \\
Number of attention heads & 4 \\
  \midrule[0.5pt]
 \multicolumn{2}{c}{\textbf{Pairformer} (\texttt{StructuralUpdate})} \\
 \midrule[0.5pt]
Dimension of single embeddings & 128 \\
Dimension of pair embeddings & 128 \\
Dimension of triangle multiplication hidden embeddings & 128 \\
Number of triangle attention heads & 4 \\
Dimension of pair attention embeddings & 32 \\
Transition layer expanding factor & 4 \\
Pair attention dropout rate & 0.25 \\
\bottomrule[1.2pt]
\end{tabular}
}
\label{tab:hyperparameters_traj}
\end{table*}

\paragraph{Structure Decoder.}

\begin{figure*}[h]
    \vspace{12pt}
    \centering
    \includegraphics[width=\linewidth]{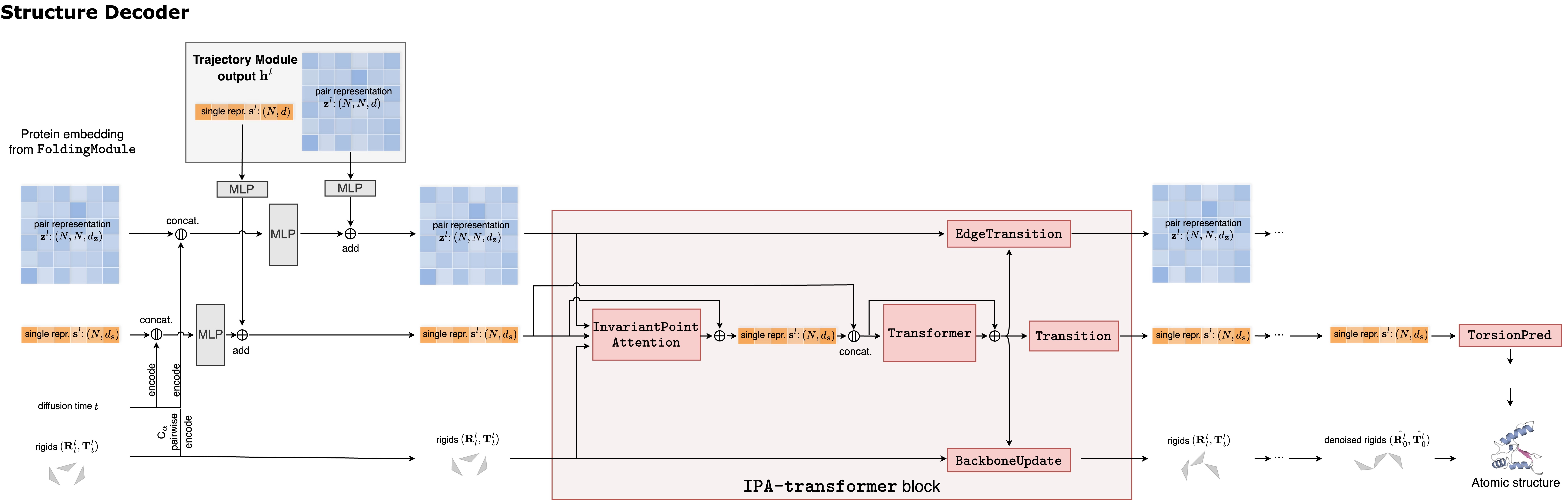}
    \caption{Architecture details of the \textit{Structure Decoder}. Single and pair embeddings from \textit{Trajectory Module} is used to update the original embeddings from \texttt{FoldingModule}. The resulting embeddings are fed into blocks of \texttt{IPA-transformer} to update single, pair embeddings and denoise rigids, SE(3) representation of protein backbone conformations. Denoised rigids together with torsion angles predicted by \texttt{TorsionPred} recovers the atomic structure of protein conformation at this frame.}
    \label{fig:arch_dec}
    \vspace{18pt}
\end{figure*}

Conformation generation, conditioned on the temporal signals from the \textit{Trajectory Module}, is performed using the model architecture from \textsc{ConfDiff}~\citep{wang2024proteinconfdiff}, As shown in Figure \ref{fig:arch_dec}, following \textsc{ConfDiff}, the inputs to the denoising model include diffusion time $t$, pairwise distance between residue rigids, and residue indices (not shown). They are encoded and concatenated with the single and pair embeddings from \texttt{FoldingModule}. In addition, the single and pair embeddings from the \textit{Trajectory Module} are projected back to the latent dimension $d_\mathbf{s}$ and $d_\mathbf{z}$, respectively, to modify the single and pair embeddings used by \textsc{ConfDiff}.

The core of the \textit{Structure Decoder} consists of multiple \texttt{IPA-transformer} blocks, which update single and pair embeddings as well as the SE(3) rigids of noisy conformations. In the final block, torsional angles are predicted by \texttt{TorsionPred} and, together with denoised rigids, to reconstruct the atomic structure of generated conformation.
The corresponding hyperparameter settings are summarized in Table~\ref{tab:hyperparameters}.

\begin{table*}[hbt!]
 \label{tab:hparam}
\centering
\caption{Hyperparameter choices of the \textit{Structure Decoder}}
\setlength{\tabcolsep}{16mm}
\begin{tabular}{ll}
\toprule[1.2pt]

\textbf{Hyperparameters} & \textbf{Values}\\ 
 \midrule[0.5pt]
 \multicolumn{2}{c}{\textbf{Neural network}} \\
 \midrule[0.5pt]
Number of IPA blocks & 4 \\
Dimension of single embedding ($d_\mathbf{s}$) & 256 \\
Dimension of pair embedding ($d_\mathbf{z}$)& 128 \\
Dimension IPA hidden embedding & 256 \\
Number of IPA attention heads & 4 \\
Number of IPA query points &  8 \\
Number of IPA value points &  12 \\
Number of transformer attention heads&4 \\
Number of transformer layers&  2 \\

  \midrule[0.5pt]
 \multicolumn{2}{c}{\textbf{SE(3)-diffusion SDE}} \\
 \midrule[0.5pt]
  Number of diffusion steps &  200 \\
  Translation scheduler &  Linear \\
  Translation $\beta_{\min}$ &  0.1 \\
  Translation $\beta_{\max}$ &  20 \\
  Rotation scheduler &  Logarithmic \\
  Rotation $\sigma_{\min}$ &  0.1 \\
  Rotation $\sigma_{\max}$ &  1.5 \\
\bottomrule[1.2pt]
\end{tabular}

\label{tab:hyperparameters}
\end{table*}

\clearpage
\subsection{Training and Inference Details}
\label{ap:training_details}

We train all \model models on the trajectories from the ATLAS training set, following the train-validation-test split of previous works~\citep{jingAlphaFoldMeetsFlow2024,wang2024proteinconfdiff,jing2024generative_mdgen}. Specifically, we exclude the training proteins longer than 384 amino acid residues, leading to 1080 training proteins.

Most components of \model models were trained from scratch, except for the \texttt{FoldingModule}, where we used frozen weights from OpenFold to extract the single and pair representations from three recycling iterations, and the \texttt{DiffusionDecoder}, which was initialized from \texttt{ConfDiff-OF-r3-MD} checkpoint provided by the authors\footnote{https://github.com/bytedance/ConfDiff}.

During each training epoch, we randomly sample stride length from $2^{0}$ to $2^{10}$ to extract sub-trajectories of length $L=9$ at varying time scales. 
With the use of causal transformers, input frames were shifted forward by one frame with a \texttt{[MASK]} token padded at the beginning of the trajectory. Combined with the use of a causal mask in temporal attention, the design ensures that each frame is trained to sample conditioned only on previous frames and the first frame is generated unconditionally using only the \texttt{[MASK]} token as input.

We trained main \model model 180 epochs ($\sim 37$ hrs) and \model[-interp] model for additional 220 epochs ($\sim 45$ hrs). Additional training hyperparameters can be found in Table \ref{tab:training_hyperparameters}.
All model training and sampling were carried out using 8 NVIDIA H100 GPUs.

\begin{table*}[hbt!]
 \label{tab:hparam_training}
\centering
\caption{Training hyperparameters}
\setlength{\tabcolsep}{16mm
}
\begin{tabular}{ll}
\toprule[1.2pt]

\textbf{Hyperparameters} & \textbf{Values}\\ 
 \midrule[0.5pt]
Batch Size  &  1 \\
Frames Num & 8 \\ 
Gradient Clip & 1.0 \\
Learning Rate   & 1$\times 10^{-4}$  \\
Optimizer  &  Adam (weight decay = 0.)\\
\bottomrule[1.2pt]
\end{tabular}

\label{tab:training_hyperparameters}
\end{table*}

\subsection{Training and Inference Cost}
\label{ap:compute_cost}

\model models in this work contains 19.6 M trainable parameters. 
All model training and sampling were carried out using 8 NVIDIA H100 GPUs with Distributed Data Parallel.
We trained the main \model model for 180 epochs ($\sim 37$ hrs) and \model[-interp] model for additional 220 epochs ($\sim 45$ hrs).

Inference cost varies with protein size and benchmark setups. To measure the potential speedups from using \model compared with classic MD simulation,  we measured the wall-clock time required to generate 100~ns trajectories (80 frames) for ATLAS proteins of varying sizes and report the average inference time per size bucket.
For comparison, we selected a representative protein from each bucket and estimated the time required to simulate 100~ns using OpenMM with implicit solvent. Both are performed on a single NVIDIA H100-80G GPU. As shown in the Table~\ref{tab:runtime_speedup}, \model provides significant speedup for 100~ns simulation, with even more pronounced acceleration for larger proteins.

\begin{table}[h!]
\centering
\fontsize{8pt}{10pt}\selectfont
\captionof{table}{Runtime comparison (in minutes) across different protein sequence lengths. Speedup is computed as MD runtime divided by \model runtime}
\label{tab:runtime}
\begin{tabular}{@{}lccccc@{}}
\toprule
\textbf{Sequence length} & (0,150) & [150,300) & [300,450) & [450,600) & [600,724] \\
\midrule
ConfRover & 6.99 & 7.53 & 10.92 & 15.88 & 20.83 \\
MD         & 104.54 & 207.92 & 386.69 & 651.29 & 1099.13 \\
\midrule
\textbf{Speedup} & \textbf{14.95$\times$} & \textbf{27.61$\times$} & \textbf{35.41$\times$} & \textbf{41.01$\times$} & \textbf{52.77$\times$} \\
\bottomrule
\end{tabular}
\label{tab:runtime_speedup}
\end{table}

For our large-scale experimental benchmarks, \textit{multi-start} trajectory simulation (2,700+ trajectories) took 8 hours and 30 minutes, while time-independent sampling (250 conformations per protein) took 3 hours and 20 minutes.

\clearpage

\newpage
\section{Additional Experimental Results}
\label{ap:exp_all}

\subsection{Trajectory Simulation: \textit{multi-start}}
\label{ap:multi-start}

\paragraph{Benchmark Curation.}
In \textit{multi-start}, we sample short trajectories from varying starting point while ensuring the generation within the scope of the reference trajectory. For example, we select frame index of 1000, 3000, 5000, and 7000 as starting frames for stride $S=128/256$, resulting in 12 test trajectories from triplicates; frame index 1000, 3000, 5000 as starting frames for stride $S=512$, and frame index 1000 for stride $S=1024$ to avoid exceeding total of 10000 frames. This provided us 2,706 different starting conditions from 82 proteins from the ATLAS test set for evaluation.

\paragraph{PCA Projection.} Following previous works~\citep{jingAlphaFoldMeetsFlow2024,wang2024proteinconfdiff}, we project the C$\alpha$ coordinates of proteins into a reduced PCA space to focus on the principal dimensions that best capture the structural variations observed in MD simulations. Briefly, for each protein, conformations from triplicate MD simulations in ATLAS are all aligned to the reference conformation (input structure for simulations). The coordinates of each C$\alpha$ atoms are then flattened and used to fit a per-protein PCA model. For all subsequent analyses, sampled conformation are aligned to the reference structure before computing their PCA projections. 

\paragraph{Additional Results.}

\begin{figure}[b]
    \centering
    \includegraphics[width=0.8\linewidth]{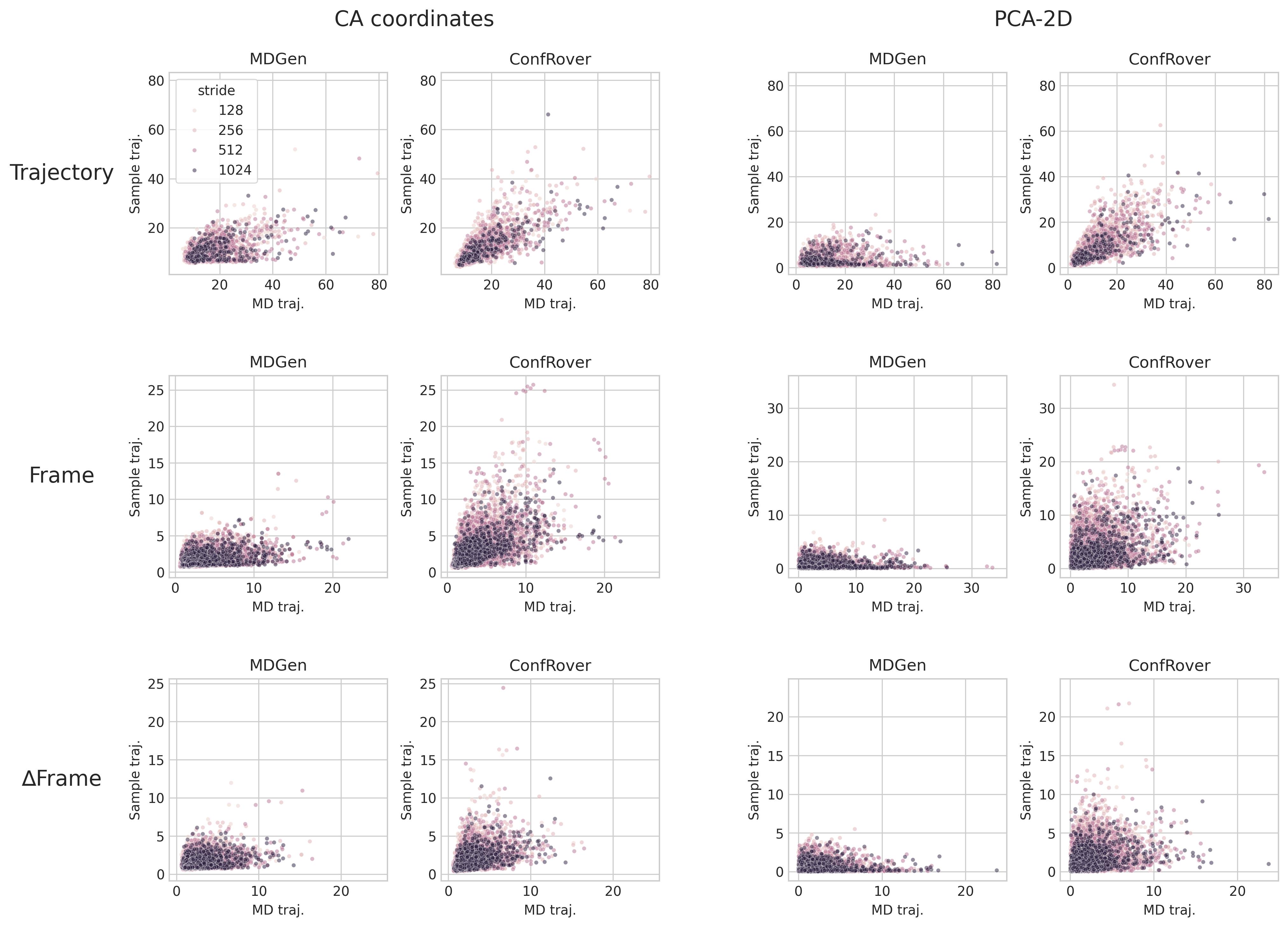}
    \caption{Scatterplots of conformation changes in sample trajectories versus those in the reference trajectories, measured by trajectory-level conformation changes, frame-level conformation changes, and next-frame difference ($\Delta$Frame), measured by the RMSD of alpha carbons (unit: Å) or $L^2$ distance in the projected PCA space. \textsc{MDGen} tends underestimate the magnitude of conformation changes while \model generate samples at similar level as the MD reference. The exact match of measured conformation changes is not possible due to stochastic sampling in both MD simulation and generative models.}
    \label{fig:fwd_vary_corr_scatter}
\end{figure}

Here we also include the scatterplot of Pearson correlation in Figure~\ref{fig:fwd_vary_corr_scatter} and additional metrics from the \textit{multi-start} experiments at different strides in Table~\ref{tab:fwd_vary_metrics} (one experimental included).  Across different strides, \model models consistently outperforms \textsc{MDGen} in recovering the correct level of dynamics.

\begin{table*}[h]
\centering
\footnotesize
\caption{Additional metrics from the \textit{multi-start} benchmark. Results for different strides are shown in separate blocks. The better score in each block is highlighted in \textbf{bold} (excluding diversity). One inference run per model is used for this comparison.}
\label{tab:fwd_vary_metrics}
{\fontsize{8pt}{9pt}\selectfont
\begin{tabular}{lcccccccc}
\toprule
 & Diversity & \multicolumn{3}{c}{MAE on PCA-2D ($\downarrow$)} & \multicolumn{3}{c}{MAE on CA coordinates ($\downarrow$)} & Quality \\
  \cmidrule(lr){2-2} \cmidrule(lr){3-5} \cmidrule(lr){6-8} \cmidrule{9-9}
 & \specialcell{Pairwise\\RMSD}     & Trajectory & Frame & \specialcell{Frame\\Next} & Trajectory & Frame & \specialcell{Frame\\Next} & \specialcell{PepBond\\Break \%} ($\downarrow$) \\
\midrule
\multicolumn{9}{l}{Stride=128} \\
\midrule
\textsc{MDGen}  & 1.26 & 6.53 & 1.28 & 0.91 & 5.02 & 0.96 & 0.71 & 27.9 \\
\model          & 1.63 & \textbf{3.10} & \textbf{1.10} & \textbf{0.74} & \textbf{3.83} & \textbf{0.81} & \textbf{0.64} & \textbf{17.3} \\
\midrule
\multicolumn{9}{l}{Stride=256} \\
\midrule
\textsc{MDGen} & 1.34 & 7.66 & 1.54 & 1.06 & 6.00 & 1.14 & 0.85 & 27.9 \\
\model & 1.78 & \textbf{3.91} & \textbf{1.28} & \textbf{0.87} & \textbf{4.60} & \textbf{0.91} & \textbf{0.75} & \textbf{16.6} \\
\midrule
\multicolumn{9}{l}{Stride=512} \\
\midrule
\textsc{MDGen} & 1.40 & 9.12 & 1.94 & 1.25 & 7.27 & 1.41 & 1.01 & 28.0 \\
\model & 1.89 & \textbf{4.84} & \textbf{1.53} & \textbf{1.01} & \textbf{5.66} & \textbf{1.07} & \textbf{0.89} & \textbf{16.7} \\
\midrule
\multicolumn{9}{l}{Stride=1024} \\
\midrule
\textsc{MDGen} & 1.51 & 11.48 & 2.62 & 1.55 & 9.04 & 1.80 & 1.24 & 28.1 \\
\model & 2.04 & \textbf{6.75} & \textbf{1.89} & \textbf{1.25} & \textbf{7.39} & \textbf{1.26} & \textbf{1.09} & \textbf{16.7} \\
\bottomrule
\end{tabular}
}
\end{table*}

\clearpage

\paragraph{Additional Visualization of Trajectory.}

\begin{figure}[h]
    \centering
    \includegraphics[width=0.98\linewidth]{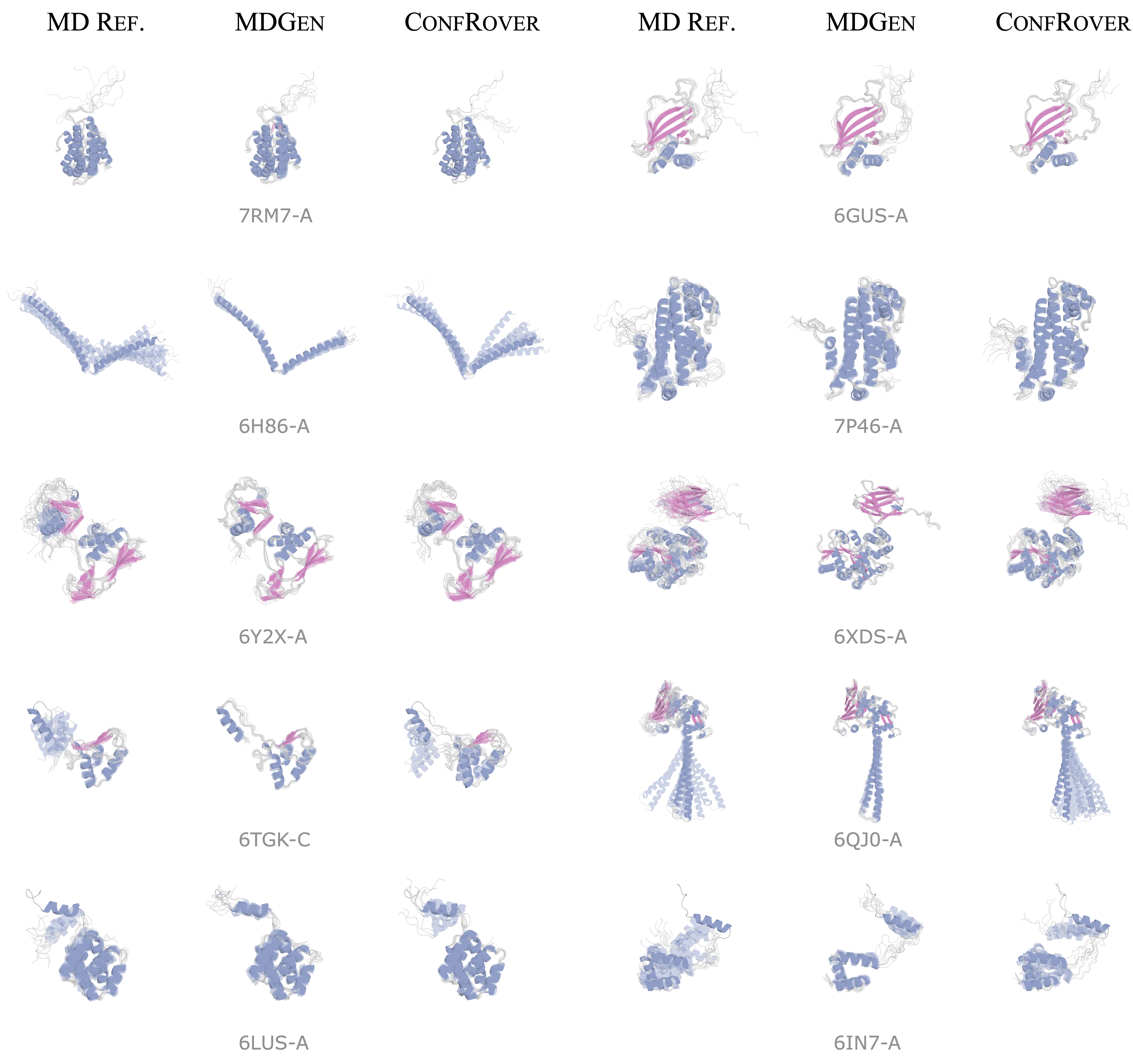}
    \caption{Visualization of 10 trajectories randomly selected from the \textit{Multi-start} benchmark. Trajectory conformations are colored by their secondary structures and superposed to show the dynamic ensemble. \textsc{MDGen} exhibits primarily local motions while \model better reflects the motions observed in MD reference.
    }
    \label{fig:fwd_vary_cases_rnd}
\end{figure}

We additional unfiltered examples (randomly selected) for visual comparison of conformations generated by different models in the \textit{multi-start} experiments, as shown in Figure~\ref{fig:fwd_vary_cases_rnd}.

\clearpage
\newpage

\subsection{Trajectory Simulation: 100~ns Long Trajectory Simulation on ATLAS}
\label{ap:fwd_100ns}

\paragraph{Details on Evaluation Metrics.}

Conformational state recovery is evaluated by comparing the distribution of model-generated and reference conformations in a PCA space. Same as in the \textit{multi-start} benchmark, each conformation is projected into the PCA space parameterized by the 3D coordinates of C$\alpha$ atoms.
To compare distributions, each principal component is discretized into 10 evenly spaced bins. After projecting the conformations into this space, we count their occurrences in each bin and compute the distribution similarity using Jensen–Shannon Distance (JSD). We also binarize the occupancy counts to compute precision, recall, and F1-score—evaluating whether sampled conformations fall within known states, following prior work~\citep{luStr2StrScorebasedFramework2024,wang2024proteinconfdiff,zhengPredictingEquilibriumDistributions2024aDiG}.

Dynamic mode recovery is assessed using time-lagged independent component analysis (tICA) applied separately to reference and generated trajectories across varying lag times. After fitting, we extract tICA coefficients for each C$\alpha$ atom and compute Pearson correlations between the per-residue contributions to the leading components, evaluating alignment of dynamic modes.

\begin{figure}[b]
    \centering
    \includegraphics[width=0.8\linewidth]{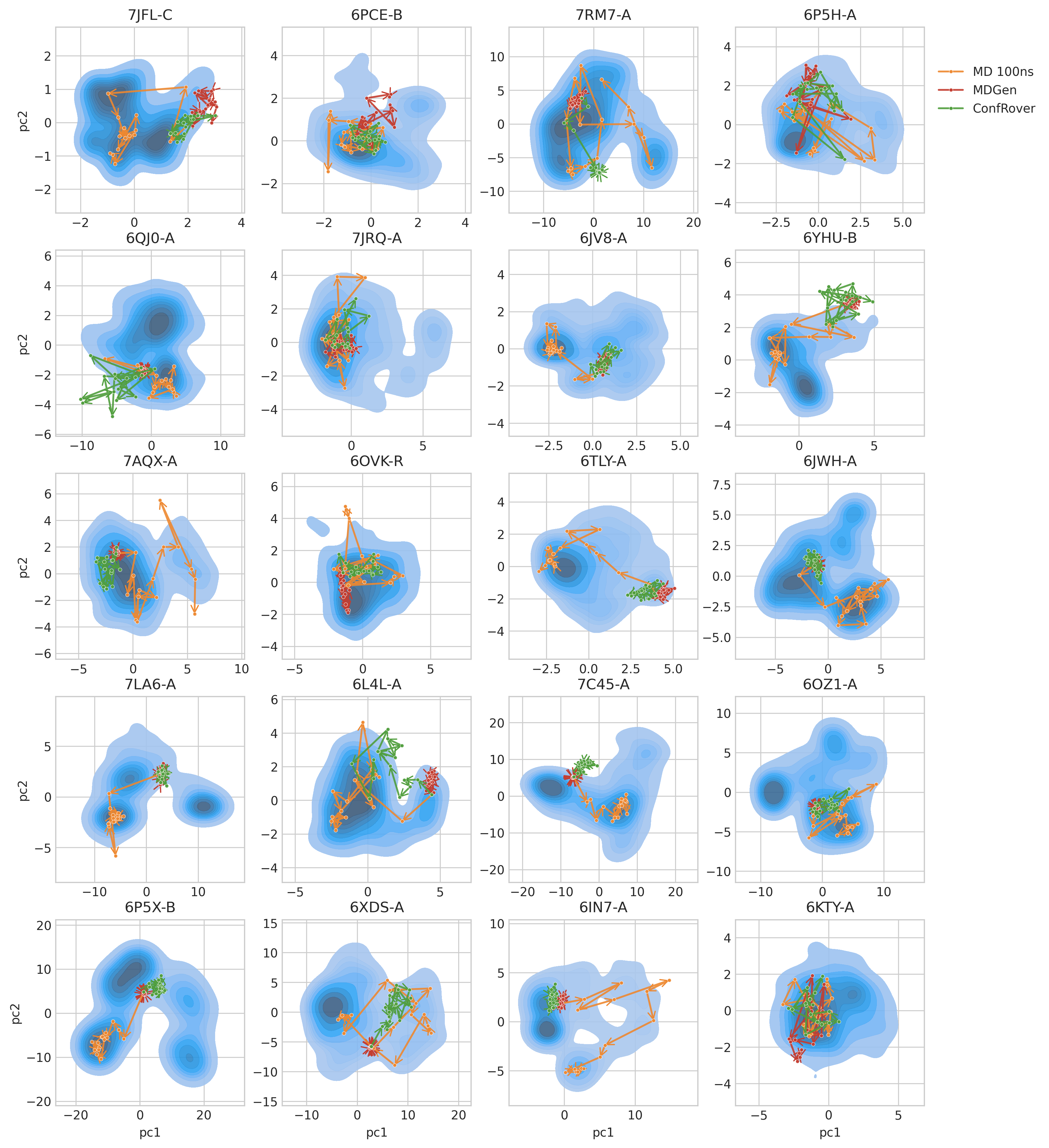}
    \caption{Visualization the ATLAS-100ns trajectories from 20 randomly selected cases. The blue background indicates the density of the ground-truth conformation distribution from MD reference. \model shows improved conformation state recovery in several cases (e.g., 7JRQ-A, 6YHU-B, 7AQX-A, 6L4L-A, 6OZ1-A, etc), sampling more diverse conformations. Yet, the gap between the oracle \textsc{MD 100ns} and deep learning models is evident in some cases (e.g., 7JFL-C, 6TLY-A, 6JWH-A, etc)}
    \label{fig:fwd_100ns_rnd_case}
\end{figure}

\paragraph{Additional Visualizations.} We additionally provide unfiltered (randomly selected) PCA plots in Figure~\ref{fig:fwd_100ns_rnd_case}.

\clearpage

\subsection{Time-independent Conformation Sampling}
\label{ap:dist}

We follow the evaluation protocol of \citet{ye2024proteinbench} to assess time-independent conformation sampling on the ATLAS test set. For each protein, 250 independent samples are generated. Since \textsc{MDGen} does not support time-independent sampling, we approximate its performance using samples from its 100-ns trajectory, serving as a sequential-sampling baseline.
The performance of state-of-the-art models, \textsc{AlphaFlow} and \textsc{ConfDiff}, is taken from Table 10 of \citet{ye2024proteinbench}, using their best-performing variants: \textsc{AlphaFlow-MD} and \textsc{ConfDiff-Open-MD}. Full results are shown in Table~\ref{tab:dist_full}.

\begin{table}[!htp]\footnotesize
  \centering\setlength{\tabcolsep}{2pt}
    \caption{Performance on time-independent conformation generation on ATLAS. A total of 250 conformations were sampled for each protein, and the mean and standard deviation of metrics are computed from five independent runs. The best performance is highlighted in \textbf{bold}, and the second-best is \underline{underlined}.\model[traj] and \textsc{MDGen} are trained to exclusive generate trajectories. \textsc{MDGen} does not support time-independent sampling and the metrics are evaluated on sequential sampling result.}
    \label{tab:dist_full}
    
    \resizebox{\textwidth}{!}{%
    \begin{tabular}{lccccccccc}
    \toprule
     & \multicolumn{2}{c}{Diversity} & \multicolumn{3}{c}{Flexibility: \textit{Pearson} $r$ on} & \multicolumn{4}{c}{Distributional accuracy} \\
     \cmidrule(lr){2-3}  \cmidrule(lr){4-6} \cmidrule(lr){7-10}
     & \specialcell{Pairwise\\RMSD} & RMSF & \specialcell{Pairwise\\RMSD ↑ } & \specialcell{Global\\RMSF ↑ }  & \specialcell{Per target\\RMSF ↑ } & \specialcell{RMWD ↓} & \specialcell{MD PCA\\$\mathcal{W}_2$ ↓} & \specialcell{Joint\\PCA $\mathcal{W}_2$↓}  & \specialcell{PC sim\\$>$ 0.5 \%↑ } \\
    \midrule
    \textsc{AlphaFlow} & \meanstd{2.85}{0.05} & \meanstd{1.63}{0.01} & \textbf{\meanstd{0.56}{0.06}} & \textbf{\meanstd{0.66}{0.04}} & \textbf{\meanstd{0.85}{0.01}} & \textbf{\meanstd{2.62}{0.03}} & \meanstd{1.52}{0.05} & \meanstd{2.26}{0.03} & \underline{\meanstd{39.52}{3.22}} \\
    \textsc{ConfDiff} & \meanstd{3.59}{0.02} & \meanstd{2.18}{0.02} & \underline{\meanstd{0.54}{0.00}} & \underline{\meanstd{0.65}{0.00}} & \textbf{\meanstd{0.85}{0.00}} & \meanstd{2.70}{0.01} & \underline{\meanstd{1.44}{0.00}} & \textbf{\meanstd{2.22}{0.04}} & \textbf{\meanstd{41.00}{1.12}} \\

    \midrule
    \textsc{MDGen} & \meanstd{1.34}{0.05} & \meanstd{0.77}{0.00} & \meanstd{0.47}{0.04} & \meanstd{0.50}{0.01} & \meanstd{0.72}{0.02} & \meanstd{2.78}{0.04} & \meanstd{1.86}{0.03} & \meanstd{2.44}{0.04} & \meanstd{10.24}{3.18} \\
    \model[traj] & \meanstd{3.19}{0.04} & \meanstd{1.74}{0.00} & \meanstd{0.48}{0.00} & \meanstd{0.62}{0.01} & \meanstd{0.84}{0.01} & \meanstd{2.85}{0.02} & \textbf{\meanstd{1.43}{0.01}} & \meanstd{2.30}{0.01} & \meanstd{37.08}{2.83} \\

    \model & \meanstd{3.68}{0.04} & \meanstd{2.25}{0.01} & \meanstd{0.51}{0.01} & \meanstd{0.64}{0.01} & \textbf{\meanstd{0.85}{0.00}} & \underline{\meanstd{2.66}{0.02}} & \meanstd{1.47}{0.03} & \underline{\meanstd{2.23}{0.04}} & \meanstd{38.28}{4.44} \\
    \bottomrule
     & \multicolumn{4}{c}{Ensemble observables} & \multicolumn{2}{c}{Quality}  \\
      \cmidrule(lr){2-5}  \cmidrule(lr){6-7}
     & \specialcell{Weak\\contacts $J$ ↑} & \specialcell{Transient\\contacts $J$↑} & \specialcell{Exposed \\residue $J$ ↑} & \specialcell{Exposed MI \\matrix $\rho$ ↑} & \specialcell{CA clash\\\% ↓} & \specialcell{PepBond\\break \% ↓} &  & &  \\
    \midrule
    \textsc{AlphaFlow} & \underline{\meanstd{0.62}{0.00}} & \textbf{\meanstd{0.41}{0.00}} & \textbf{\meanstd{0.69}{0.01}} & \textbf{\meanstd{0.35}{0.01}} & \textbf{\meanstd{0.00}{0.00}} & \meanstd{22.00}{0.19} \\
    \textsc{ConfDiff} & \textbf{\meanstd{0.64}{0.00}} & \underline{\meanstd{0.40}{0.00}} & \underline{\meanstd{0.67}{0.00}} & \underline{\meanstd{0.33}{0.00}} & \meanstd{0.50}{0.00} & \textbf{\meanstd{6.20}{0.00}} \\
    \midrule
    \textsc{MDGen} & \meanstd{0.51}{0.01} & \meanstd{0.28}{0.01} & \meanstd{0.57}{0.01} & \meanstd{0.26}{0.01} & \underline{\meanstd{0.24}{0.05}} & \meanstd{30.84}{1.69} \\
    \model[traj] & \meanstd{0.53}{0.01} & \meanstd{0.36}{0.00} & \meanstd{0.58}{0.01} & \meanstd{0.27}{0.00} & \meanstd{0.32}{0.04} & \underline{\meanstd{11.78}{0.24}} \\

    \model & \underline{\meanstd{0.62}{0.01}} & \meanstd{0.37}{0.01} & \meanstd{0.66}{0.01} & \meanstd{0.32}{0.01} & \meanstd{0.50}{0.00} & \meanstd{19.18}{0.08} \\
    \bottomrule
    \end{tabular}
    }%
\end{table}%

\clearpage

\subsection{Conformation Interpolation}
\label{ap:interp}

In \textit{conformation interpolation} experiment, we selected trajectories from \textit{multi-start}. These trajectories exhibit sufficient conformation changes (e.g., RMSD between the start and end frames $> 4$ Å) and clear interpolation path in the PCA space. The list of selected cases are in Table~\ref{tab:interp_cases}.

\begin{figure}[htbp]
  \centering

  \begin{subfigure}[t]{0.48\textwidth}
    \centering
    \includegraphics[width=\linewidth]{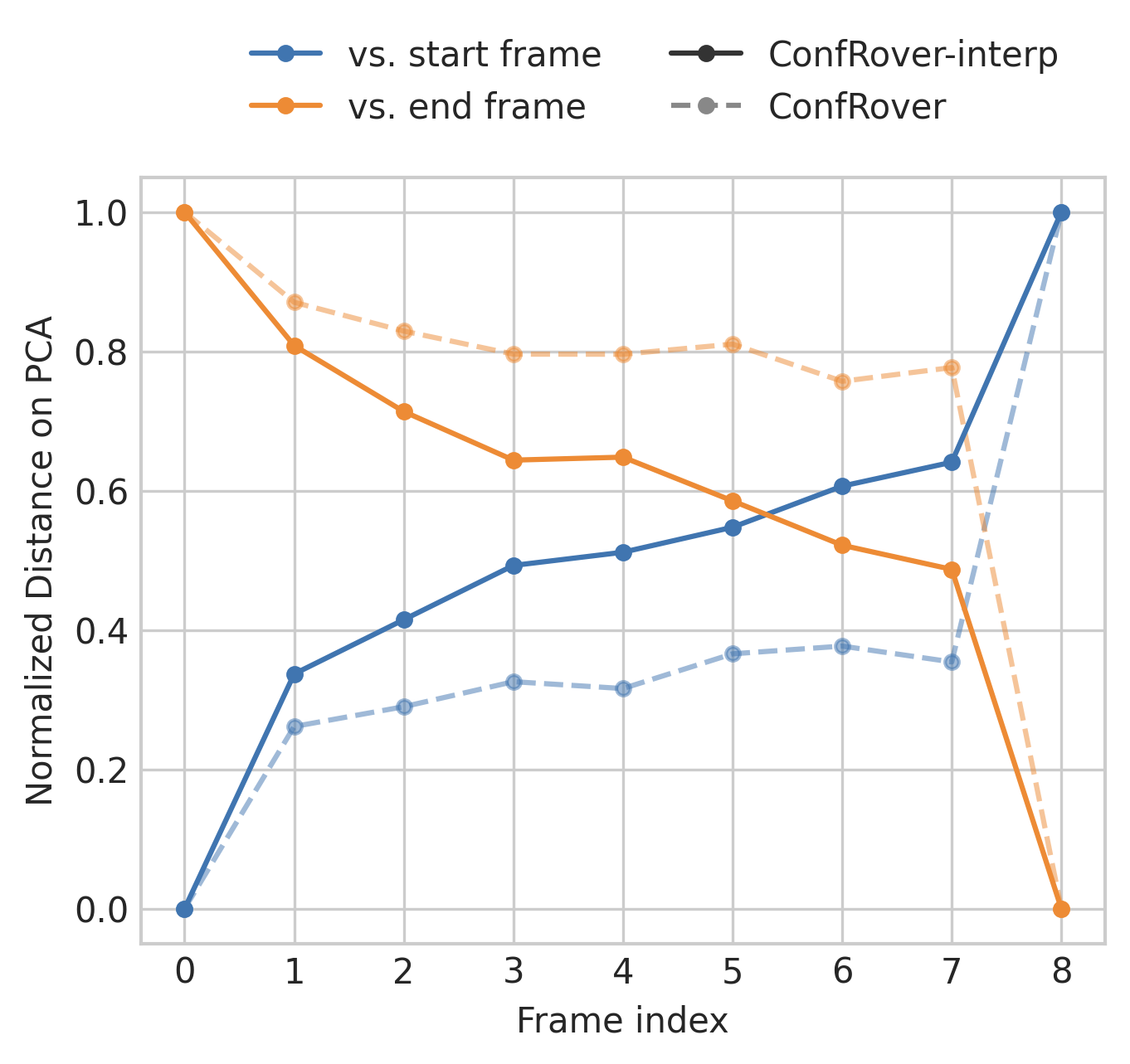}
  \end{subfigure}
  \hfill
  \begin{subfigure}[t]{0.48\textwidth}
    \centering
    \includegraphics[width=\linewidth]{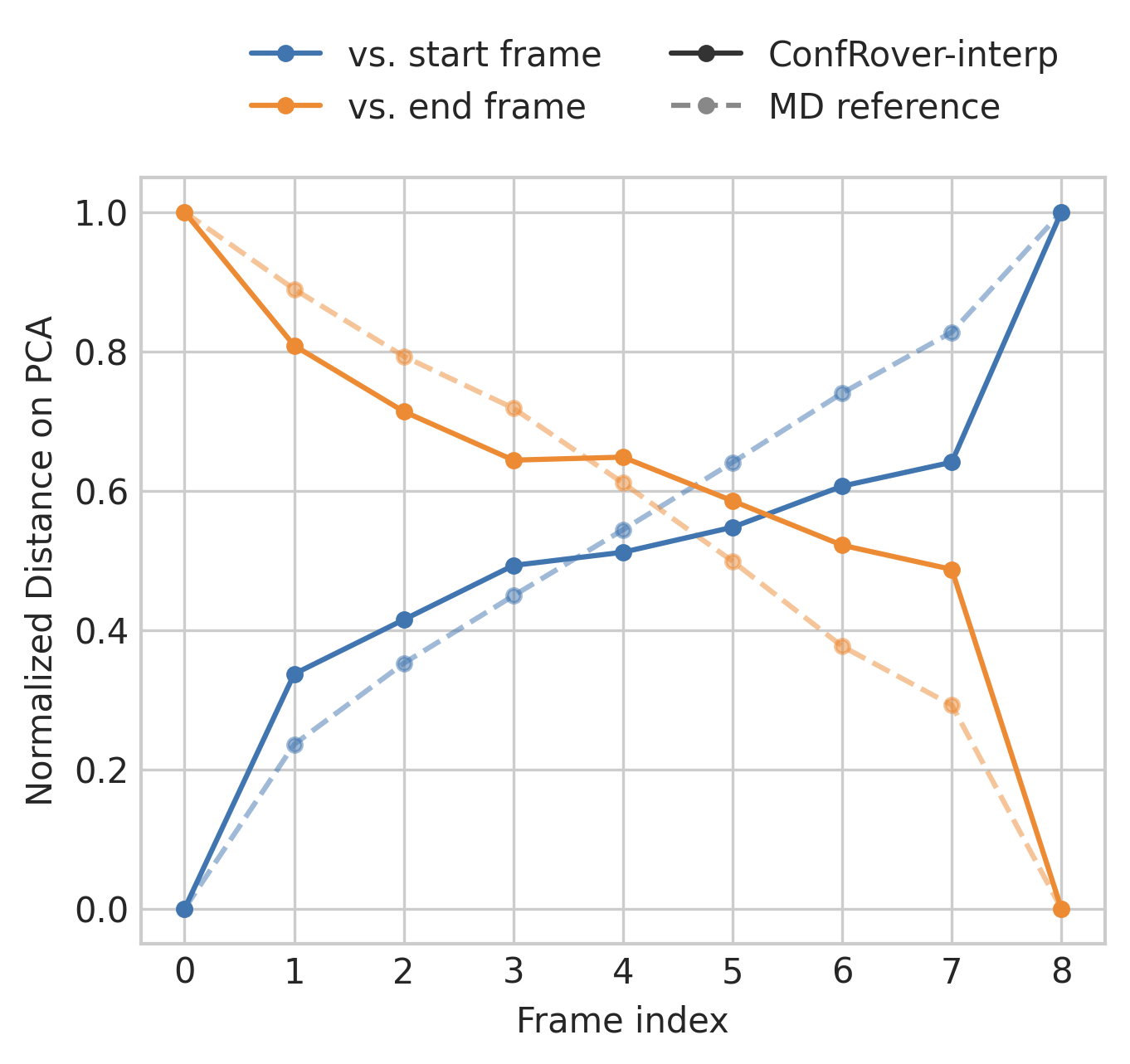}
  \end{subfigure}

  \caption{Normalized PCA distance of intermediate frames to the start and end frames, averaged over 38 cases selected from the \textit{multi-start} benchmark. [Left] a comparison between \model[interp] and \model, where \model[interp] generates smooth pathways while \model does not; [Right] a comparison between \model[interp] and reference trajectories. }
  \label{fig:interp_curves}
\end{figure}

The $L^2$ distance of generated intermediate frames to the start and end frames in the PCA spaces are reported in Figure~\ref{fig:interp_curves}. Distances are normalized by the distance between start and end frames. Similar to the results measured by C$\alpha$-RMSD, \model[interp] shows smooth interpolation between the start and end frames while \model does not. This result shows that by continue training the model on interpolation objective, \model can learn to generated interpolating trajectories conditioned on the end state.

\begin{table}[h]
    \centering
    \footnotesize
    \caption{List of 38 selected cases from \textit{multi-init} for interpolation test. Naming conventions: ``\{PDB\_ID\}\_\{Chain\_ID\}\_R\{ATLAS repeat\}F\{Starting index\}S\{Stride\}''}
    \label{tab:interp_cases}
    \vspace{6pt}
    {\fontsize{8pt}{9pt}\selectfont
    \begin{tabular}{lllll}
    \toprule
    0 & 5ZNJ-A-R2F1000S512 & 6L4L-A-R1F3000S256 & 6TGK-C-R1F5000S256 & 7JFL-C-R2F3000S256 \\
    \midrule
    1 & 6E7E-A-R3F3000S512 & 6LRD-A-R1F1000S1024 & 6TLY-A-R2F1000S256 & 7JRQ-A-R1F1000S1024 \\
    \midrule
    2 & 6GUS-A-R2F7000S256 & 6LUS-A-R2F3000S128 & 6XDS-A-R1F1000S128 & 7LA6-A-R1F1000S512 \\
    \midrule
    3 & 6H49-A-R2F1000S1024 & 6OVK-R-R2F7000S128 & 6XRX-A-R3F5000S128 & 7LP1-A-R2F5000S256 \\
    \midrule
    4 & 6H86-A-R2F1000S1024 & 6OZ1-A-R1F1000S512 & 6Y2X-A-R2F3000S512 & 7P41-D-R2F3000S256 \\
    \midrule
    5 & 6IN7-A-R3F5000S128 & 6P5H-A-R1F1000S1024 & 7AEX-A-R3F5000S256 & 7P46-A-R3F5000S256 \\
    \midrule
    6 & 6J56-A-R1F3000S256 & 6P5X-B-R1F3000S256 & 7AQX-A-R2F3000S512 & 7RM7-A-R3F3000S128 \\
    \midrule
    7 & 6JPT-A-R2F5000S512 & 6Q9C-A-R3F3000S256 & 7ASG-A-R1F7000S128 & 7S86-A-R3F5000S256 \\
    \midrule
    8 & 6JV8-A-R3F1000S256 & 6QJ0-A-R1F1000S256 & 7BWF-B-R3F3000S128 &  \\
    \midrule
    9 & 6KTY-A-R3F1000S1024 & 6RRV-A-R1F1000S256 & 7C45-A-R1F5000S512 &  \\
    \bottomrule
    \end{tabular}
    }
\end{table}

\clearpage
\newpage
\paragraph{Additional Visualizations.} Here we include additional visualization on interpolation results.

\begin{figure}[H]
    \centering
    \includegraphics[width=0.9\linewidth]{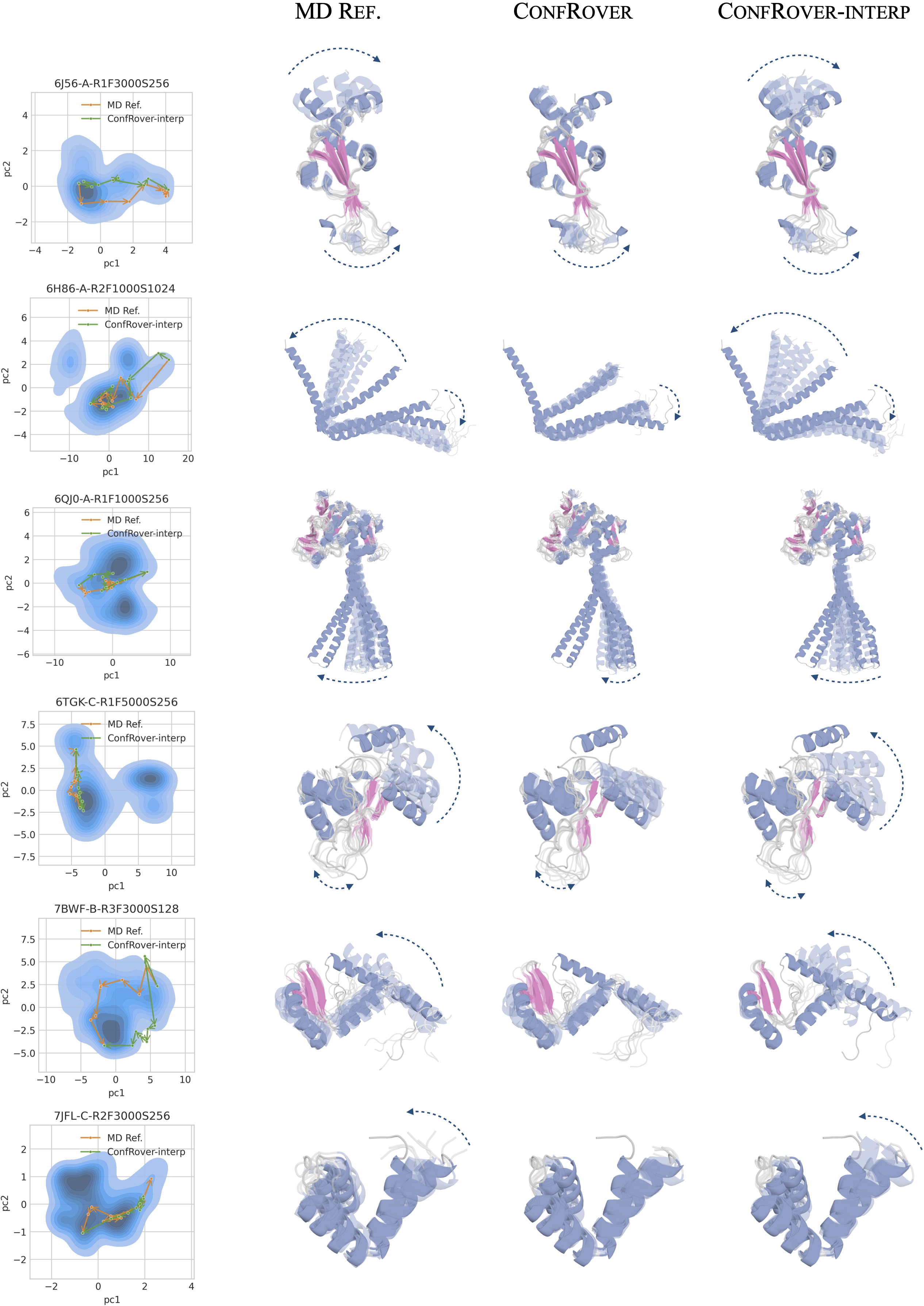}
    \caption{Example interpolations results. \model[interp] generates smooth pathways between the start and end frames, capturing the dynamics observed with the MD reference while \model does not show the correct intermediate conformations. Start and end frames are shown as solid structures; intermediate conformations are shown in fading colors. Main motions are indicated by blue dashed arrows. These examples highlight the difference between the original \model and \model[interp] that further trained on the interpolation objective. The original \model can miss key motions of the transition while \model[interp] correctly capture these motions.}
    \label{fig:interp_add_vis}
\end{figure}

\begin{figure}[H]
    \centering
    \includegraphics[width=0.9\linewidth]{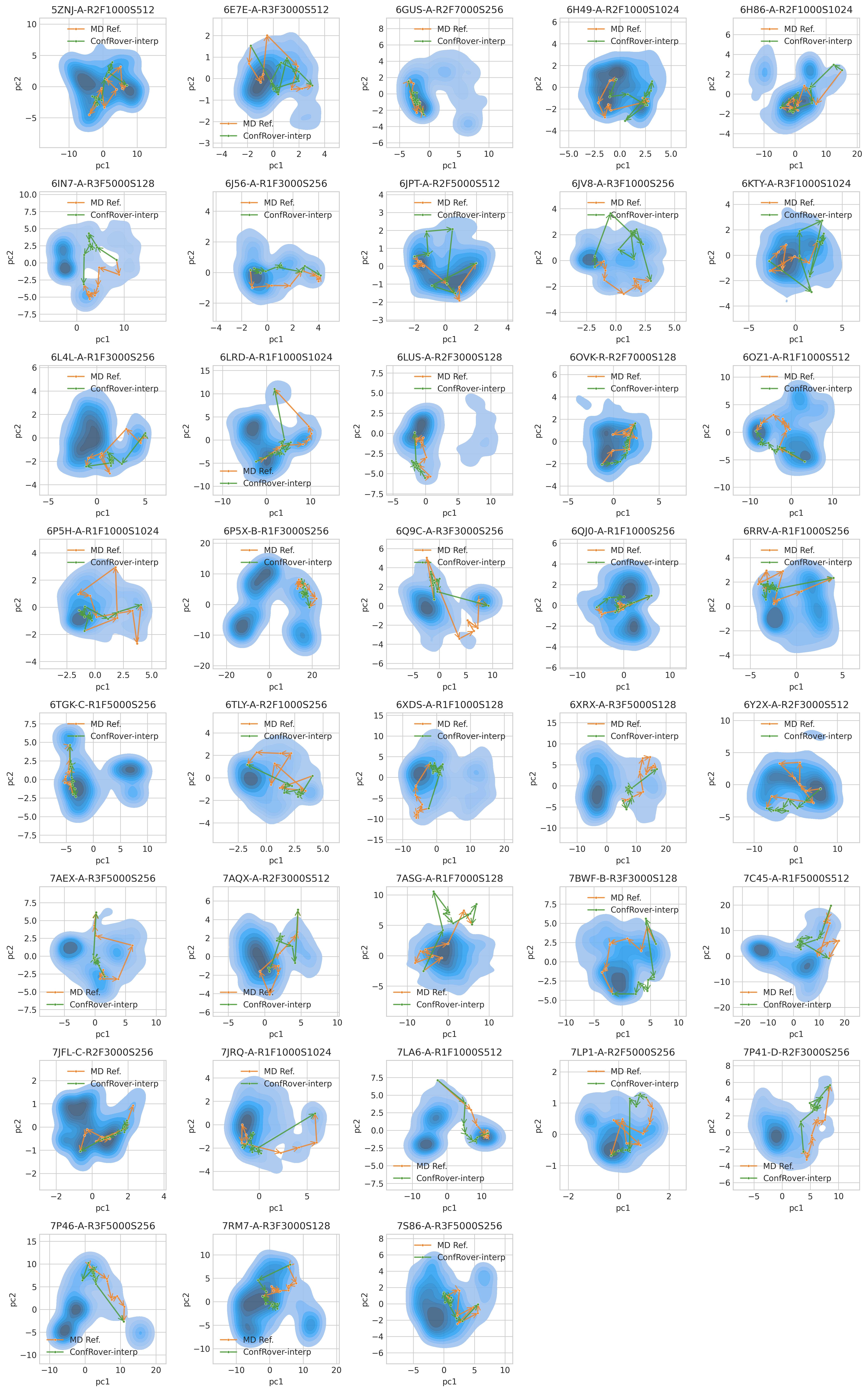}
    \caption{PCA plot of 38 selected interpolation cases. MD reference trajectories and results from \model[interp] are shown in each plot.}
    \label{fig:interp_case38.}
\end{figure}

\subsection{Extended Baseline Comparisons}
\subsubsection{Retraining MDGen for Experiment-Specific Setups}
\label{ap:mdgen_retrain}
\begin{figure}[H]
    \centering
    \includegraphics[width=0.5\linewidth]{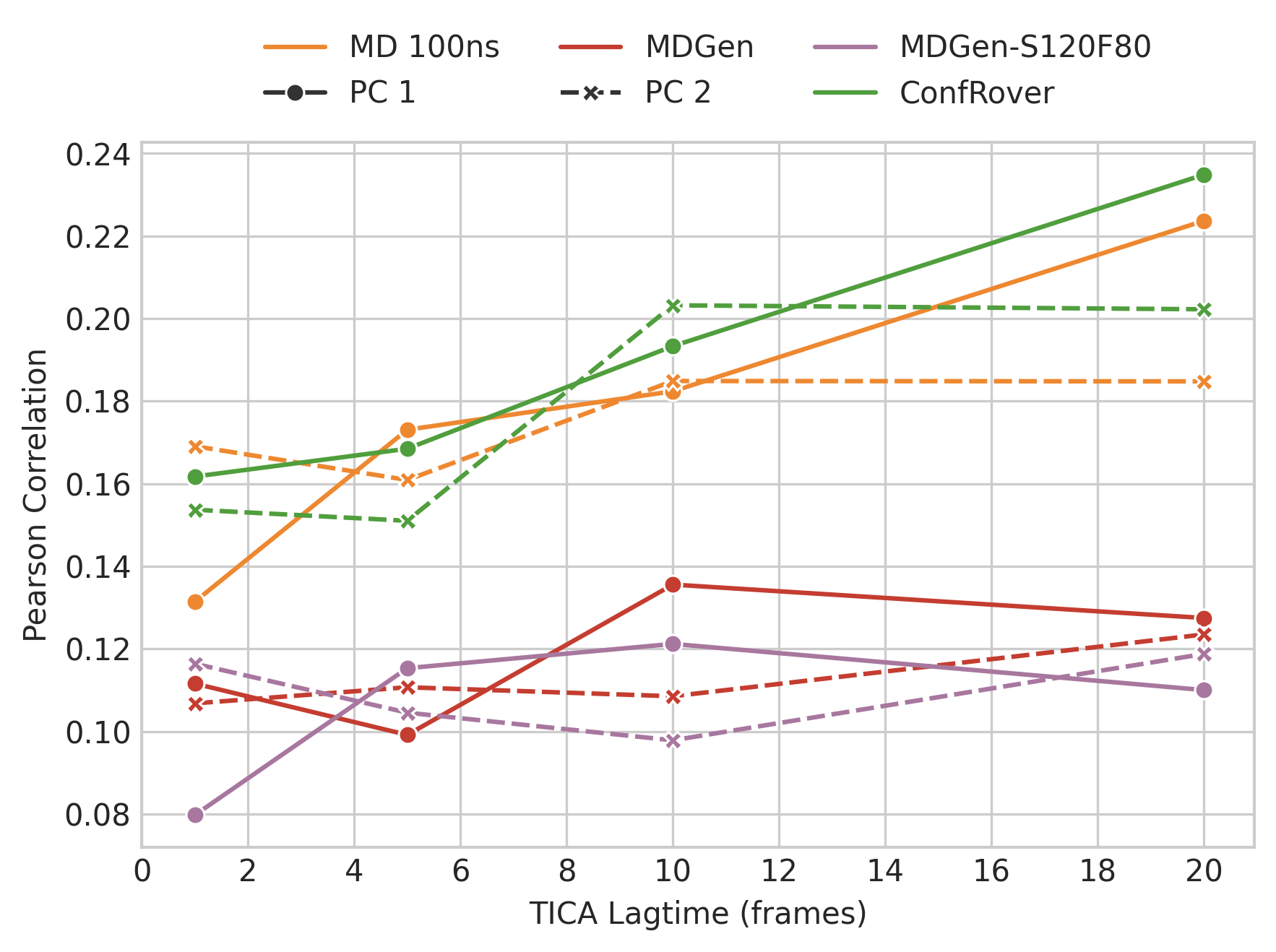}
    \caption{Compare \textsc{MDGen-S256F9} with \textsc{MDGen} from subsampling post-process. This figure shows correlations of main dynamic modes between sampled trajectory and reference trajectory. \textsc{MDGen-S120F80} is trained and sampled with stride of 120 MD snapshots and length of 80 frames.}
    \label{fig:fwd_100ns_tica_mdgen_retrain}
\end{figure}

Due to its non-autoregressive design and training on fixed-length trajectories, \textsc{MDGen} cannot directly generate sequences of varying lengths. Therefore, in our evaluation, we generate trajectories using the original settings (named \textsc{MDGen}-S\{stride\}F\{length\}) and subsample them to match the evaluation setup. However, this post-processing may introduce artifacts. To address this concern, we retrain \textsc{MDGen} under the evaluation settings and compare the results on the \textit{Multi-start} benchmark (stride~=~256) and 100 ns long trajectory simulations, as shown below in Table~\ref{tab:fwd_vary_corr_mdgen_retrain}, Table~\ref{tab:fwd_vary_metrics_mdgen_retrain}, Table~\ref{tab:fwd_100ns_state_mdgen_retrain} and Figure~\ref{fig:fwd_100ns_tica_mdgen_retrain}. Experimental results show no significant difference of performance for key metrics observed comparing \textsc{MDGen} with post-processed results and models specifically trained at the evaluation settings, suggesting no evident decrease of trajectory quality from the subsampling post-process. We use one inference run per model for this experiment.

\begin{table}[h]
    \centering
    \footnotesize
    \captionof{table}{Compare \textsc{MDGen-S256F9} with \textsc{MDGen} from subsampling post-process. Here is the table summarizing the Pearson correlations of conformation changes between sampled and reference trajectories in \textit{multi-start}. \textsc{MDGen-S256F9} is trained and sampled with stride of 256 MD snapshots and length of 9 frames. The best scores are highlighted in \textbf{bold}.}
    \label{tab:fwd_vary_corr_mdgen_retrain}
        \begin{tabular}{cccc}
            \toprule
            & \multicolumn{3}{c}{C$\alpha$ coordinates} \\
            \cmidrule{2-4}
            & Trajectory & Frame & $\Delta$Frame \\
            \midrule
            \textsc{MDGen} & 0.57 & 0.46 & 0.41 \\
            \textsc{MDGen-S256F9} & 0.56 & 0.45 & 0.38 \\
            \model & \textbf{0.77} & \textbf{0.62} & \textbf{0.53} \\
            \midrule
            & \multicolumn{3}{c}{PCA 2D} \\
            \cmidrule{2-4}
            & Trajectory & Frame & $\Delta$Frame \\
            \midrule
            \textsc{MDGen} & 0.18 & 0.13 & 0.11 \\
            \textsc{MDGen-S256F9} & 0.21 & 0.19 & 0.11 \\
            \model & \textbf{0.75} & \textbf{0.5} & \textbf{0.44} \\
            \bottomrule
        \end{tabular}
\end{table}

\begin{table*}[h]
\centering
\footnotesize
\caption{Compare \textsc{MDGen-S256F9} with \textsc{MDGen} from subsampling post-process. Here is the table summarizing additional metrics in \textit{multi-start} benchmark. \textsc{MDGen-S256F9} is trained and sampled with stride of 256 MD snapshots and length of 9 frames. The best scores are highlighted in \textbf{bold}.}
\label{tab:fwd_vary_metrics_mdgen_retrain}
\begin{tabular}{lcccccccc}
\toprule
 & Diversity & \multicolumn{3}{c}{MAE on PCA-2D ($\downarrow$)} & \multicolumn{3}{c}{MAE on CA coordinates ($\downarrow$)} & Quality \\
  \cmidrule(lr){2-2} \cmidrule(lr){3-5} \cmidrule(lr){6-8} \cmidrule{9-9}
 & \specialcell{Pairwise\\RMSD}     & Traj. & Frame & $\Delta$Frame & Traj. & Frame & $\Delta$Frame & \specialcell{PepBond\\Break \%($\downarrow$)} \\
\midrule
\textsc{MDGen} & 1.34 & 7.66 & 1.54 & 1.06 & 6.00 & 1.14 & 0.85 & 27.9 \\
\textsc{MDGen-S256F9} & 1.59 & 6.90 & 1.47 & 1.03 & 5.33 & 1.13 & 0.82 & \textbf{16.2} \\
 \model & 1.78 & \textbf{3.91} & \textbf{1.28} & \textbf{0.87} & \textbf{4.60} & \textbf{0.91} & \textbf{0.75 }& 16.6 \\
\bottomrule
\end{tabular}
\end{table*}

\begin{table}[H]
\centering
\footnotesize
\caption{Compare \textsc{MDGen-S256F9} with \textsc{MDGen} from subsampling post-process. Here is the table summarizing the state recovery performance in 100 ns long trajectory simulation. \textsc{MDGen-S120F80} is trained and sampled with stride of 120 MD snapshots and length of 80 frames. The best scores are highlighted in \textbf{bold}. \textsc{MD 100ns} is included as the oracle.}
\label{tab:fwd_100ns_state_mdgen_retrain}
\begin{tabular}{lrrr}
\toprule
 & JSD ($\downarrow$) & Recall ($\uparrow$) & F1 ($\uparrow$) \\
\midrule
\rowcolor{Highlight}
\textsc{MD 100ns} & 0.31 & 0.67 & 0.79 \\
\textsc{MDGen} & 0.56 & 0.30 & 0.44 \\
\textsc{MDGen-S120F80} & 0.57 & 0.29 & 0.42 \\
\model & \textbf{0.51} & \textbf{0.42} & \textbf{0.58} \\
\bottomrule
\end{tabular}
\end{table}

\subsubsection{Internal Reproduction of AlphaFolding}
\label{ap:alphafolding}
Available baseline models for protein trajectory generation are limited at the time of this work, we attempted to reproduce AlphaFolding~\citep{cheng20244d4ddiff}, a diffusion-based model that generate short trajectories (i.e., blocks) and can be extended to longer through iterative generation.
We followed the authors' official implementation~\footnote{https://github.com/fudan-generative-vision/dynamicPDB/tree/main/applications/4d\_diffusion}, setting the motion token count to 2 and the generation horizon (block length) to 16 frames. To improve generation efficiency, we increased the stride from the default 1 to 40, matching the setup in \textsc{MDGen}.
Training \textsc{AlphaFolding} on the full ATLAS dataset resulted in out-of-memory error on NVIDIA A100-80GB GPU. Therefore, we adopted the authors’ filtering criterion of a maximum sequence length of 256 residues. Similarly, we encountered the out-of-memory error for the five largest proteins during inference. 
The model was trained for 65K steps where we saw convergence. For evaluation, we iteratively extended the 16-frame outputs to generate 256 frames and retained the first 250 frames to match the 100~ns simulation setting.

We compared \textsc{AlphaFolding} with \textsc{MDGen} and \model on the ATLAS 100~ns simulation task, using one inference run per model. As shown in Table~\ref{tab:alphafolding_tica_corr}, both \textsc{AlphaFolding} and \model{} outperform \textsc{MDGen} in capturing the principal coordinates of the dominant dynamics, although \textsc{AlphaFolding} slightly lags behind \model. 
However, as shown in Table~\ref{tab:alphafolding_quality}, \textsc{AlphaFolding} produces lower-quality conformations, exhibiting inflated backbone and side-chain rotamer outliers as well as higher clash rates. We found more evident degradations from error accumulation (Table~\ref{tab:alphafolding_quality_frames}), likely due to its iterative block-wise extension, which conditions only on the last frame of the preceding block. In contrast, \textsc{MDGen} employs non-autoregressive attention across all frames, while \model maintains a full attention history via KV cache.
Owing to the noisy nature of its generated trajectories, \textsc{AlphaFolding} also shows inflated recall in conformational state recovery (Table~\ref{tab:alphafolding_state_coverage}), as the increased structural noise likely leads to artificially higher coverage in state space.

Although our implementation is a reproduction of \citep{cheng20244d4ddiff} and may not faithfully reflect the authors' original model, this analysis suggests the potential limitations of block-extension-based trajectory models including \textsc{AlphaFolding}.

\begin{table}[H]
\centering
\footnotesize
\caption{Pearson correlations of principal dynamic modes (PC1 and PC2) between sampled and reference trajectories, evaluated at varying lag times ($\Delta t$, in frames). The best scores are highlighted in \textbf{bold}.}
\label{tab:alphafolding_tica_corr}
\begin{tabular}{@{}llcccc@{}}
\toprule
 &  & $\Delta t = 1$ & $\Delta t = 5$ & $\Delta t = 10$ & $\Delta t = 20$\\
\midrule
PC 1 & \textsc{MDGen}         & 0.11 & 0.12 & 0.10 & 0.12 \\
PC 1 & \textsc{AlphaFolding}  & 0.15 & 0.15 & 0.16 & 0.18 \\
PC 1 & \model     & \textbf{0.19} & \textbf{0.17} & \textbf{0.19} & \textbf{0.19} \\
\midrule
PC 2 & \textsc{MDGen}         & 0.12 & 0.10 & 0.10 & 0.11 \\
PC 2 & \textsc{AlphaFolding}  & 0.18 & 0.15 & 0.16 & \textbf{0.20} \\
PC 2 & \model     & \textbf{0.19} & \textbf{0.17} & \textbf{0.18} & 0.17 \\
\bottomrule
\end{tabular}
\end{table}

\begin{table}[H]
\centering
\footnotesize
\caption{Conformation geometric quality evaluated using MolProbity. Un-relaxed conformations are used for evaluation. The best scores are highlighted in \textbf{bold}.}
\label{tab:alphafolding_quality}
\begin{tabular}{@{}lccccccc@{}}
\toprule
&
\specialcell{Ramachandran\\outliers \% ($\downarrow$)} & 
\specialcell{Rotamer\\outliers \% ($\downarrow$)} & 
\specialcell{Clash\\score ($\downarrow$)} & 
\specialcell{RMS\\bonds ($\downarrow$)} & 
\specialcell{RMS\\angles ($\downarrow$)} & 
\specialcell{MolProbity\\score ($\downarrow$)} \\

\midrule
\textsc{AlphaFolding} & 2.91  & 15.37 & 151.35 & 0.07 & 4.76 & 3.94 \\
\textsc{MDGen}        & 1.87  & \textbf{2.85}  & 128.08 & \textbf{0.04} & \textbf{3.14} & 3.28 \\
\model    & \textbf{1.75}  & 3.30  & \textbf{76.89}  & 0.05 & 3.69 & \textbf{3.00} \\
\bottomrule
\end{tabular}
\end{table}

\begin{table}[H]
\centering
\footnotesize
\caption{Average Ramachandran outliers across trajectory frame ranges. Compared with \textsc{MDGen} and \model, \textsc{AlphaFolding} exhibits an increased level of backbone outliers when simulating longer trajectories.}
\label{tab:alphafolding_quality_frames}
\begin{tabular}{@{}lccccc@{}}
\toprule
\textbf{Frame range} & (0,15] & (15,31] & (31,47] & (47,63] & (63,79] \\
\midrule
\textsc{AlphaFolding} & 1.77 & 2.78 & 3.21 & 3.34 & 3.57 \\
\textsc{MDGen}        & 1.59 & 1.85 & 1.96 & 1.95 & 2.05 \\
\model    & 1.34 & 1.61 & 1.77 & 1.93 & 2.08 \\
\bottomrule
\end{tabular}
\end{table}

\begin{table}[H]
\centering
\footnotesize
\caption{Recovery of conformational states in the ATLAS 100~ns simulation experiment. Although \textsc{AlphaFolding} shows the highest coverage, this may due to lower sample quality that artificially inflates diversity.}
\label{tab:alphafolding_state_coverage}
\begin{tabular}{@{}lccc@{}}
\toprule
& JS-Dist~$\downarrow$ & Recall~$\uparrow$ & F1~$\uparrow$ \\
\midrule
\textsc{AlphaFolding} & 0.47 & 0.51 & 0.65 \\
\textsc{MDGen}        & 0.55 & 0.29 & 0.43 \\
\model    & 0.51 & 0.42 & 0.58 \\
\bottomrule
\end{tabular}
\end{table}

\clearpage
\newpage
\subsection{Extension to Masked Sequence Modeling}
\begin{wrapfigure}{r}{0.35\textwidth} 
    \centering
    \vspace{-28pt}
    \includegraphics[width=0.8\linewidth]{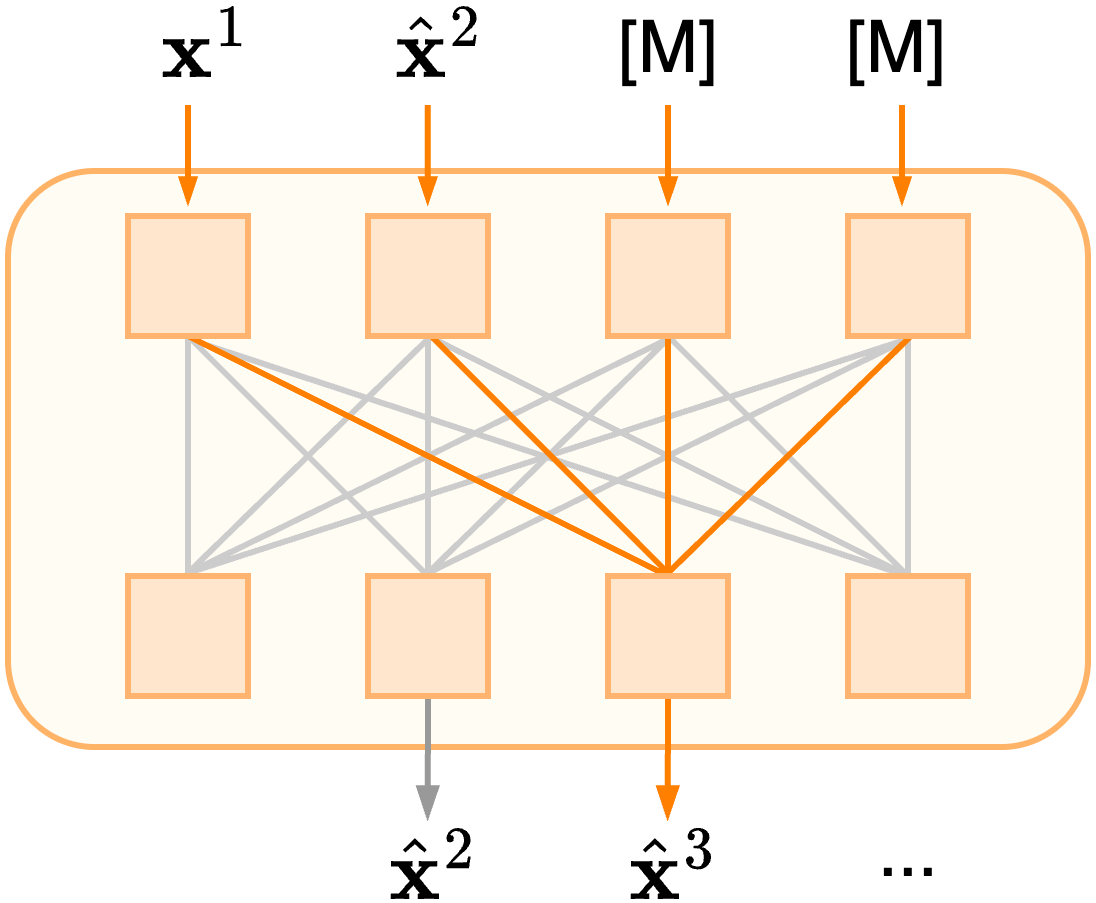}
    \caption{Autoregressive generation with \textit{masked language modeling}. $\mathbf{x}^1$ is the initial frame for simulation. Masked input frames are iteratively replaced by the generated frames. Attention activations for the current predicting frame $\hat{\mathbf{x}}^3$ are highlighted in orange. }
    \label{fig:causal_vs_mask}
    \vspace{-18pt}
\end{wrapfigure}

While the causal transformers used in \model enables efficient autoregressive training and generation, proposed framework can be extended to other sequence modeling paradigm, such as \textit{masked sequence modeling}.
Similar to~\citet{li2024autoregressivemar}, sequence models trained with bidirectional attention and masked sequence modeling can perform autoregressive generation by iteratively predicting frames in a specified order. At each iteration, the generated frame replaces its corresponding mask and joins the input sequence for the next iteration. We trained a variant \model[mask] following this approach. Specifically, we replaced causal attention with bidirectional attention in the transformer and implemented a scheduled masking strategy for training, linearly increasing the mask rate from 45\% to 88\%.

\noindent\textit{ConfRover-Mask learns conformation changes in trajectories but lacks inference flexibility.} 
 We evaluated \model[mask] on the \textit{multi-start} benchmark. As shown in Table~\ref{tab:fwd_vary_corr_masked}, \model[mask] outperforms the baseline, though it performs slightly worse than the \model (causal) model. These results demonstrate the flexibility of our framework to extend to a broader range of sequence modeling approaches. However, masked sequence modeling presents several practical limitations. Similar to non-autoregressive model, \model[mask] lacks the flexibility to generate trajectories of variable lengths. We observed structural degradation when inferring with lengths differ significantly from the training setup, such as setting $L=1$ for time-independent conformation sampling. As a result, this approach is not well-suited for generating long trajectories without resorting to workarounds such as sliding window prediction with window sizes matching the training window size. 
While training improvements, such as varying training window sizes, may help address these issues, we leave such exploration to future work.

Overall, the comparison between \model (causal) and \model[mask] highlights the versatility of causal sequence modeling, which can be more readily adapted to diverse generation scenarios.

\begin{table}[h]
    
    \centering
    \captionof{table}{Results of \model[mask] on the Pearson correlations of conformation changes between sampled and reference trajectories in \textit{multi-start}. The better scores are highlighted in \textbf{bold}.}
    \label{tab:fwd_vary_corr_masked}
    {\fontsize{8pt}{9pt}\selectfont
    \begin{tabular}{cccc}
        \toprule
        & \multicolumn{3}{c}{C$\alpha$ coordinates} \\
        \cmidrule{2-4}
        & Traj. & Frame & $\Delta$Frame \\
        \midrule
        \textsc{MDGen} & 0.55 & 0.45 & 0.40 \\
        \model & \textbf{0.77} & \textbf{0.63} & \textbf{0.53} \\
        \model[mask] & 0.71 & 0.60 & 0.49 \\
        \midrule
        & \multicolumn{3}{c}{PCA 2D} \\
        \cmidrule{2-4}
        & Traj. & Frame & $\Delta$Frame \\
        \midrule
        \textsc{MDGen} & 0.16 & 0.11 & 0.10 \\
        \model & \textbf{0.75} & \textbf{0.50} & \textbf{0.43} \\
        \model[mask] & 0.73 & 0.49 & 0.41 \\
        \bottomrule
    \end{tabular}
    }
\end{table}

\end{document}